\documentclass[11pt]{article}

\usepackage[final]{acl}

\usepackage{times}
\usepackage{latexsym}

\usepackage[T1]{fontenc}

\usepackage[utf8]{inputenc}

\usepackage{microtype}

\usepackage{inconsolata}

\usepackage{graphicx}

\newcommand{\CAI}{\textsc{CAI-dLLM}}
\usepackage[ruled,vlined]{algorithm2e}
\usepackage{amsmath}
\usepackage{amssymb}
\usepackage{booktabs}
\usepackage{tabularx}

\title{\textsc{CAI-dLLM}: Convergence Aware Inference for Diffusion Language Models
}

\author{
  Farhana Amin \\
  Virginia Tech \\
  \texttt{afarhana@vt.edu} \And
  Sabiha Afroz \\
  Virginia Tech \\
  \texttt{sabihaafroz@vt.edu} \And
  Dimitrios S. Nikolopoulos \\
  Virginia Tech \\
  \texttt{dsn@vt.edu}
}

\usepackage{multirow}
\usepackage{array}
\usepackage{caption}
\usepackage{xcolor}
\usepackage{colortbl}
\usepackage{amsmath}
\usepackage{placeins}
\definecolor{caihighlight}{RGB}{255, 248, 220}

\usepackage{amsmath}
\usepackage{amssymb}
\usepackage{makecell}
\usepackage[ruled,vlined]{algorithm2e}
\usepackage[most]{tcolorbox}

\usepackage[most]{tcolorbox}
\usepackage{xcolor}

\definecolor{snippetblue}{HTML}{4FA8CF}
\definecolor{snippetbg}{HTML}{EAF4FB}
\definecolor{snippetborder}{HTML}{5AA9E6}

\newcounter{snippetbox}
\renewcommand{\thesnippetbox}{\arabic{snippetbox}}

\newtcolorbox{snippetboxenv}[2][]{
  enhanced,
  breakable,
  colback=snippetbg,
  colframe=snippetborder,
  boxrule=0.8pt,
  arc=0pt,
  outer arc=0pt,
  left=3pt,
  right=3pt,
  top=8pt,
  bottom=8pt,
  fonttitle=\bfseries,
  fontupper=\small,
  coltitle=white,
  title={Box~\thesnippetbox: #2},
  colbacktitle=snippetblue,
  attach boxed title to top left={
    xshift=6pt,
    yshift=-7pt
  },
  boxed title style={
    colback=snippetblue,
    colframe=snippetblue,
    boxrule=0pt,
    arc=0pt,
    outer arc=0pt,
    left=7pt,
    right=7pt,
    top=4pt,
    bottom=4pt
  },
  before upper={\vspace{1em}},
  #1
}

\newcommand{\snippetbox}[3]{%
  \refstepcounter{snippetbox}%
  \begin{snippetboxenv}{#1}%
  #2%
  \label{#3}%
  \end{snippetboxenv}%
}

\newcommand{\SA}[1]{\noindent\textcolor{violet}{\bf $\blacksquare$ Sabiha: #1}}

\begin{document}
\maketitle

\begin{abstract}
Diffusion language models can generate many tokens in parallel, but they still require repeated denoising steps during inference. This makes generation costly, especially when the model continues to recompute tokens that are already stable. To address these limitations, we propose \textsc{CAI-dLLM}, a training-free inference method that uses first-step confidence to guide denoising and reduce inference time. Specifically, \textsc{CAI-dLLM} commits easy tokens earlier, allocates more denoising steps to harder tokens, and adjusts decoding schedules across output blocks. As it relies only on first-step confidence signals, it does not require retraining, extra predictors, or weight updates. We evaluate \textsc{CAI-dLLM} on LLaDA-8B-Instruct and Dream-7B-Instruct across math, code, reasoning, commonsense, and long-context (Dream-7B). \textsc{CAI-dLLM} achieves up to 18.2$\times$  wall clock inference speedup on LLaDA GSM8K while improving accuracy (77.41\% 
vs.\ 76.27\%), and up to 13.1$\times$ on Dream 
HumanEval at higher pass@1 than no-cache (48.17\% 
vs.\ 46.95\%). On harder reasoning tasks, speedups reach $44.8\times$, with a largest accuracy drop of 4.4 points, while energy consumption is reduced by up to 95.3\%.
\end{abstract}

\section{Introduction}
\label{sec:intro}

Large language models are commonly built on autoregressive 
generation, producing text from left to right one token at 
a time~\citep{vaswani2017attention, brown2020language, 
grattafiori2024llama}. This creates a sequential dependency 
during inference: each token depends on all previous tokens, 
so long outputs require many sequential model calls and 
generation is slow.

Masked diffusion language models offer a parallel alternative to autoregressive decoding by refining masked tokens over multiple denoising steps and updating many positions at once~\citep{nie2025llada,ye2025dream}. This can improve throughput, but each denoising step still runs a full transformer forward pass over the sequence, including tokens that have already stabilized. Recent methods reduce the cost of each step through KV caching~\citep{wu2025fastdllm,ma2025dkvcache,liu2025dllmcache} or early layer token skipping~\citep{zhu2026esdllm}. However, they still use the same denoising budget for every token. Easy tokens continue to receive unnecessary steps, while hard tokens receive no extra refinement.

Our key observation is that the confidence from the first 
denoising step provides a free, reliable estimate of token 
difficulty. Tokens with high first-step confidence 
stabilize after only a few steps, while tokens with low 
confidence require more refinement. Since this signal is 
produced during the required first forward pass, it adds 
no extra cost.

Based on this observation, we propose \textsc{CAI-dLLM}, 
a novel training-free inference method that uses first-step 
confidence to assign each token its own step budget, set 
block level commit schedules, and detect and exit 
low yield late denoising. Easy tokens are committed early 
and hard tokens receive more steps, directly addressing 
the limitation that existing methods cannot vary 
per token computation. Our contributions are as follows.
\begin{itemize}

\item \textbf{First-step confidence as a token level 
control signal.} We show that confidence at the first 
denoising step predicts which tokens stabilize early 
and reuse this signal throughout decoding at no extra 
cost, without re-evaluating at every step as in 
Fast-dLLM~\citep{wu2025fastdllm}.

\item \textbf{Block level adaptive threshold scheduling.} 
We introduce a block end threshold $\theta_e^{(b)}$ that 
decreases across output blocks, reflecting that later 
blocks converge faster given more committed context.

\item \textbf{Confidence gated token budgets and 
grinding phase detection.} We assign each token a 
maximum denoising budget $B_i \in \{8, 32, 64\}$ based 
on first-step confidence and position, and introduce a 
low yield detector that stops denoising when newly 
committed tokens remain scarce for $K$ consecutive steps.

\end{itemize}
\section{Background and Motivation}
\label{sec:background}

Masked diffusion language models generate text through an iterative denoising process~\citep{austin2021structured,lou2023discrete,sahoo2024simple}.
Given an input prompt, the model initializes a set of masked positions and progressively replaces them with predicted tokens.
At each denoising step, the model predicts a vocabulary distribution for every masked position, and a decoding rule determines which tokens are committed.
Unlike autoregressive models, which generate tokens sequentially, masked diffusion models can update multiple positions in a single forward pass.
However, inference remains costly because generation still requires many denoising steps.
As a result, the total cost depends on both the output length and the number of denoising iterations.

\subsection{Denoising Computation is Uneven}
\label{sec:motivation_redundancy}

A fixed decoding policy applies the same computation rule to all tokens and output blocks.
This assumption is inefficient because denoising does not progress uniformly across the sequence.
We analyze this behavior using LLaDA-8B-Instruct on GSM8K with 256 generated tokens divided into four 64-token blocks~\citep{cobbe2021gsm8k}.

We measure the change in hidden states between adjacent denoising steps using relative L2 drift:
\begin{equation}
D^{(t)} =
\frac{
\|\mathbf{h}^{(t)}-\mathbf{h}^{(t-1)}\|_2
}{
\|\mathbf{h}^{(t-1)}\|_2
}.
\label{eq:relative_drift}
\end{equation}
Here, $\mathbf{h}^{(t)} \in \mathbb{R}^{L \times d}$ denotes the hidden state at denoising step $t$.
We average this value over token positions.

Our analysis shows that early layers change substantially less than deeper layers:
\begin{equation}
\small
\begin{aligned}
D_{\mathrm{early}}(L0\text{--}L8)
&\approx 2\text{--}3\%,\\
D_{\mathrm{deep}}(L17\text{--}L31)
&\approx 6\text{--}14\%.
\end{aligned}
\label{eq:layer_drift_pattern}
\end{equation}

\begin{figure*}[ht]
  \centering
  \includegraphics[width=0.95\textwidth]{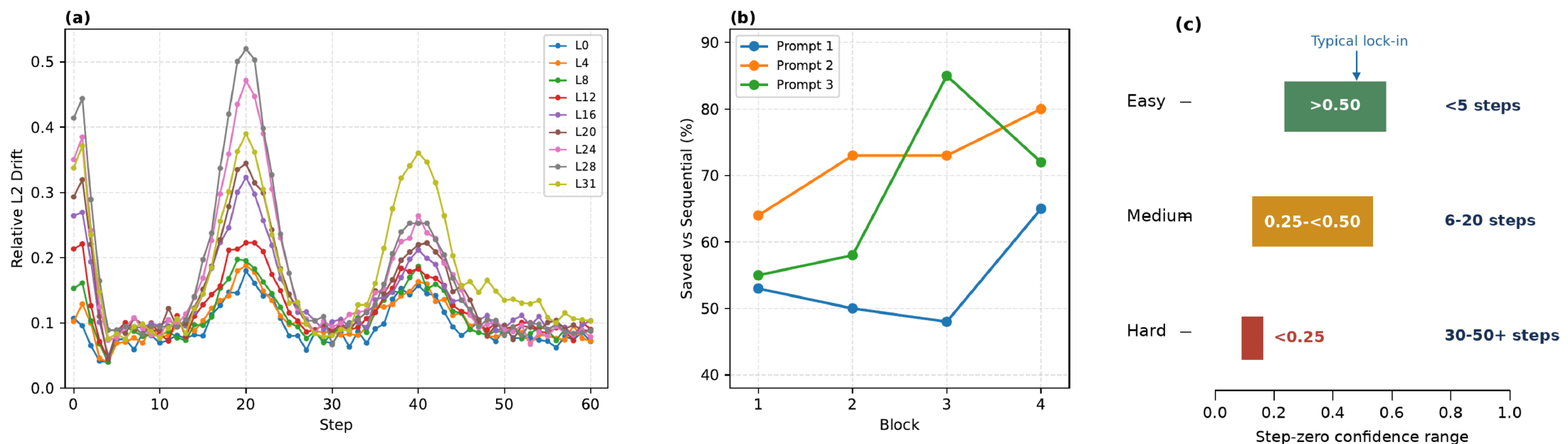}
  \caption{
  Denoising work is uneven across layers, blocks, and tokens.
  (a) Hidden state drift varies across transformer layers.
  (b) Saved steps increase in later blocks as earlier context becomes available.
  (c) First-step confidence separates easy tokens with $c_i^{(0)} > 0.50$, medium tokens with $0.25 \leq c_i^{(0)} \leq 0.50$, and hard tokens with $c_i^{(0)} < 0.25$.
  }
  \label{fig:bg_motivation_combined}
\end{figure*}

We also observe that first-step confidence separates tokens by convergence difficulty:
\begin{equation}
\small
\begin{aligned}
c_i^{(0)} > 0.50
&\Rightarrow t_i^{*} \approx 5,\\
c_i^{(0)} < 0.25
&\Rightarrow t_i^{*} \approx 30\text{--}50.
\end{aligned}
\label{eq:token_stability_pattern}
\end{equation}
Here, $t_i^{*}$ denotes the first denoising step at which token $i$ becomes stable.
We define a token as stable when its top prediction remains unchanged for three consecutive steps. These trends are measured across 1,319 GSM8K test prompts.
Figure~\ref{fig:bg_motivation_combined} summarizes the same behavior across layers, output blocks, and tokens. Additional motivation analysis is provided in Appendix~\ref{app:background_motivation}.

\subsection{First-Step Confidence as a Control Signal}
\label{sec:motivation_confidence}

The key question is whether token difficulty can be estimated early.
We use the confidence from the first denoising step as this signal.
For a masked position $i$, we define

\begin{equation}
    c_i^{(0)} =
    \max_{v \in \mathcal{V}}
    p_{\theta}(x_i = v \mid x^{(0)}).
    \label{eq:first_conf}
\end{equation}

This value is already computed during standard decoding.
It does not require another model, another forward pass, or retraining.
As shown in Figure~\ref{fig:bg_motivation_combined}(c), tokens with $c_i^{(0)} > 0.50$ usually stabilize early, while tokens with $c_i^{(0)} < 0.25$ often need many more denoising steps.
We therefore use $c_i^{(0)}$ to guide token commitment and step allocation.

\section{Related Work}
\label{sec:related}

\paragraph{Autoregressive language models.}
Autoregressive models generate text one token at a 
time~\citep{vaswani2017attention,brown2020language,grattafiori2024llama}. 
KV caching~\citep{kwon2023vllm} reduces repeated attention 
computation but does not remove the sequential bottleneck, 
motivating parallel diffusion decoding~\citep{lee2018iterative}.

\paragraph{Diffusion language models.}
LLaDA~\citep{nie2025llada} and Dream~\citep{ye2025dream} show 
that masked diffusion models match autoregressive models on 
reasoning, math, and code, but every token receives the same 
fixed number of steps regardless of convergence speed. This 
uniform treatment is the core inefficiency \textsc{CAI-dLLM} addresses.

\paragraph{Caching for diffusion language models.}
Several methods reduce inference cost by reusing KV 
activations across denoising steps: dKV-Cache~\citep{ma2025dkvcache} 
delays reuse until tokens stabilize, dLLM-Cache~\citep{liu2025dllmcache} 
applies separate rules for prompt and response tokens, 
Fast-dLLM~\citep{wu2025fastdllm} commits tokens when confidence 
exceeds a fixed threshold and introduces DualCache, which 
we use as the Fast-dLLM representative baseline. 
FlashDLM~\citep{hu2025flashdlm} combines caching with autoregressive 
guidance and Sparse dLLM~\citep{song2025sparsedllm} prunes 
less useful cache entries. These methods reduce repeated computation through cache reuse or cache sparsity.

\paragraph{Parallel and hybrid decoding.}
D2F (Discrete Diffusion Forcing;~\citealt{wang2025d2f}) mixes 
autoregressive and diffusion decoding to make KV reuse easier, but this reduces the 
parallelism that makes diffusion models fast in the first place. 
Spiffy~\citep{agrawal2026spiffy} adapts speculative 
decoding~\citep{leviathan2023fast} to diffusion 
models, but the draft-and-verify overhead adds 
complexity and extra computation at each step. Both methods 
require changes to the decoding structure itself. \textsc{CAI-dLLM} 
keeps the original model and decoding process unchanged, making 
it simpler to apply and compatible with existing caching methods.

\paragraph{Adaptive computation and token skipping.}
ES-dLLM skips low importance token computation in early layers~\citep{zhu2026esdllm}.
This is effective, but it still runs the full denoising loop.
ES-dLLM reduces computation inside each denoising step, while \textsc{CAI-dLLM} reduces the number of denoising steps assigned to each token.
\section{Method}
\label{sec:method}

\begin{figure*}[t]
  \centering
  \includegraphics[width=0.8\textwidth]{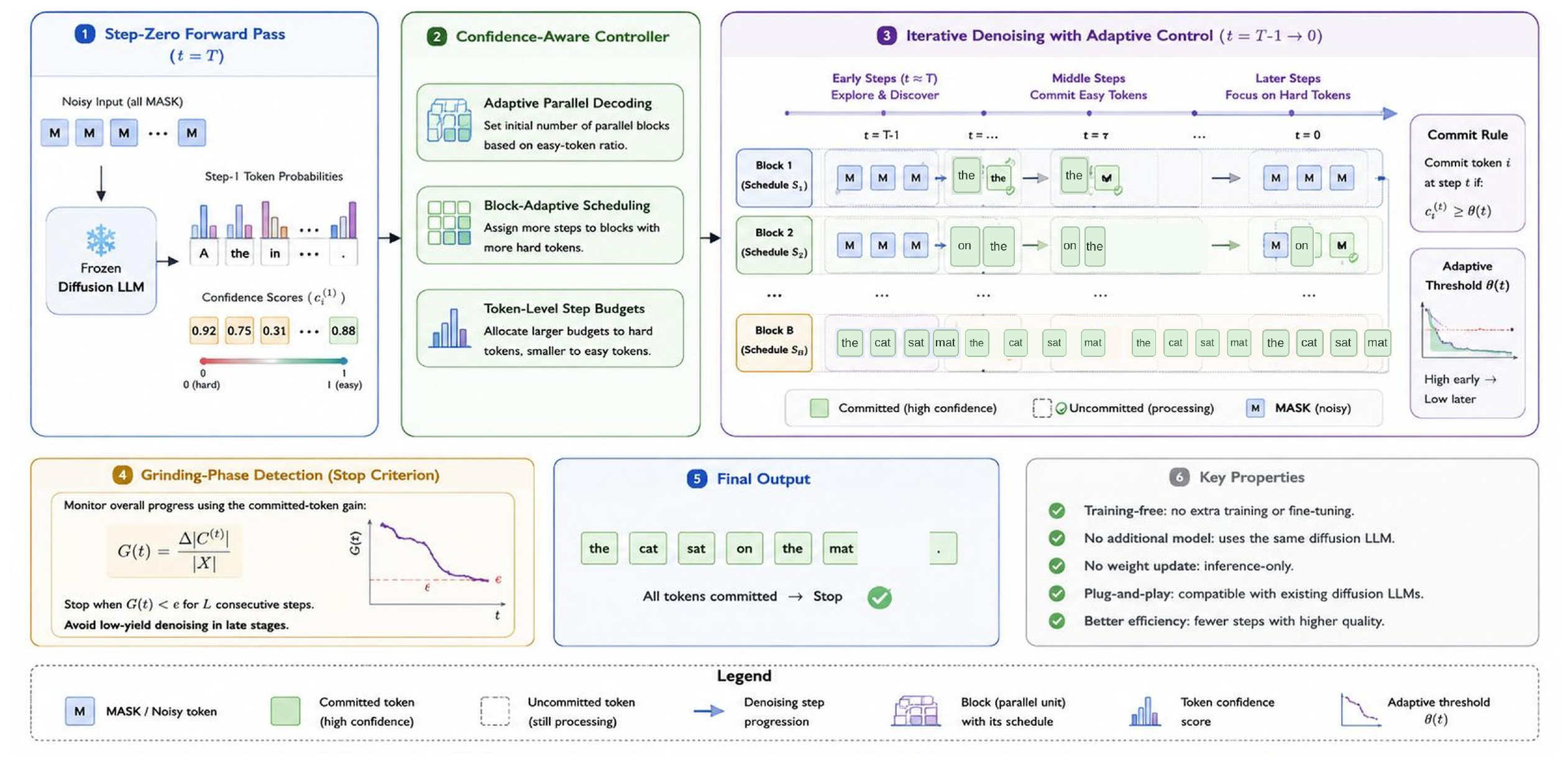}
  \caption{Overview of \textsc{CAI-dLLM}.
  \textcircled{1}~The method runs a step-zero forward 
  pass to obtain token confidence scores.
  \textcircled{2}~These scores drive a confidence-aware 
  controller that guides adaptive parallel decoding, 
  block-adaptive scheduling, and token-level step budgets.
  \textcircled{3}~During iterative denoising, easy tokens 
  are committed early while hard tokens receive more steps.
  \textcircled{4}~A grinding-phase detector stops 
  low-yield late denoising.
  The method is training-free and does not change model 
  weights.}
  \label{fig:cai_method_overview}
\end{figure*}

\textsc{CAI-dLLM} is a training-free inference method that uses first-step confidence to control masked diffusion decoding.
Figure~\ref{fig:cai_method_overview} summarizes the full pipeline: the model first runs the required step-zero (first step) forward pass to obtain token confidence scores, then a confidence aware controller uses these scores to guide adaptive parallel decoding, block level threshold schedules, token budgets, and low yield early exit.
The method builds on confidence based parallel decoding~\citep{wu2025fastdllm}, but replaces the fixed commit rule with a lightweight controller that commits easy tokens early, gives hard tokens more steps, and stops late denoising when few new tokens are being resolved.

\subsection{Diffusion Decoding Setup}
\label{sec:method_setup}

Let $\mathbf{x}=(x_1,\ldots,x_L)$ be a sequence of $L$ tokens.
At denoising step $t$, the model keeps a partially decoded sequence $\mathbf{x}^{(t)}$.
Some positions contain generated tokens.
Other positions still contain the masked tokens.
Let $\mathcal{M}^{(t)}$ be the set of masked positions at step $t$.
For each masked position, the model predicts a distribution over the vocabulary. Box~\ref{snip:token_confidence} defines token confidence. 

\snippetbox{Token Confidence}{
\label{snip:token_confidence}
For each masked token $i \in \mathcal{M}^{(t)}$, the diffusion language model predicts

\[
p_{\theta}(x_i \mid \mathbf{x}^{(t)}),
\]

where $\mathcal{V}$ is the vocabulary.
We define the confidence of token $i$ at step $t$ as

\[
c_i^{(t)}
=
\max_{v \in \mathcal{V}}
p_{\theta}(x_i=v \mid \mathbf{x}^{(t)}).
\]

A high value of $c_i^{(t)}$ means that the model has a strong current prediction for token $i$.
A low value means that the token is still uncertain.
}{snip:confidence}

\subsection{First-Step Confidence}
\label{sec:step_zero}

The first denoising step is required in every diffusion 
decoding process. We use this step to compute a 
difficulty signal for each token at no extra cost, 
since no additional forward pass is needed beyond what 
standard decoding already performs (Box~\ref{snip:step_zero}).

\snippetbox{First-Step Confidence Signal}{
\label{snip:first_step_confidence}
For each token $i$, we store the confidence from the first denoising step:
\[
s_i = c_i^{(0)}.
\]
The value $s_i$ is used as an early estimate of token difficulty.
Tokens with high $s_i$ are treated as easier.
Tokens with low $s_i$ are treated as harder.
This signal adds no extra model call because the first forward pass is already part of normal decoding.
}{snip:step_zero}

As established in Section~\ref{sec:motivation_confidence} 
and shown in Figure~\ref{fig:bg_motivation_combined}(c), 
our first observation is that tokens with high 
first-step confidence ($c_i^{(0)} > 0.50$) stabilize 
within 5 steps in nearly all cases, and tokens with 
low confidence ($c_i^{(0)} < 0.25$) continue changing 
for 30 steps or more. This pattern is consistent across 
prompts and tasks, which means $c_i^{(0)}$ is a 
reliable signal for controlling token commitment and 
step allocation.

\snippetbox{Adaptive Threshold Schedule}{
Let $T$ be the maximum number of denoising steps for one block.
Let $\theta_{\mathrm{s}}$ and $\theta_{\mathrm{e}}$ be the start and end thresholds.
Let $w$ be the warmup fraction.
\[
u(t)=\frac{t-wT}{T-wT}
\]
\[
\alpha=\frac{\theta_{\mathrm{s}}-\theta_{\mathrm{e}}}{2}
\]
\[
\theta(t)=
\begin{cases}
\theta_{\mathrm{s}}, & t < wT, \\[3pt]
\theta_{\mathrm{e}}+\alpha\bigl(1+\cos(\pi u(t))\bigr),
& t \geq wT.
\end{cases}
\]
A token is committed when
\[
c_i^{(t)} \geq \theta(t).
\]
We use $\theta_{\mathrm{s}}=0.90$, $\theta_{\mathrm{e}}=0.70$, and $w=0.15$.
The schedule stays strict early and relaxes later.
}{snip:apd}

We use $s_i$ as defined in Box~\ref{snip:step_zero} 
to assign each token a step budget, set commit 
thresholds, and schedule denoising across blocks. 
Storing $s_i$ and the per-token budgets adds 
$\mathcal{O}(L)$ memory and $\mathcal{O}(L)$ work per 
step, negligible relative to the $\mathcal{O}(L^2 d)$ 
transformer forward pass. We do not measure this overhead in isolation; however, the end-to-end throughput results in Tables~\ref{tab:dream} and~\ref{tab:llada} confirm that it does not reduce the observed speedups.


\subsection{Block Adaptive Scheduling}
\label{sec:block_schedule}
Long outputs are generated in blocks. Our second observation is that later blocks benefit from more committed context produced by earlier blocks, so they converge faster and can use a lower commit threshold without accuracy loss.
Figure~\ref{fig:bg_motivation_combined}(b) shows that steps saved increase across the four output blocks.
Savings are lowest in Block~1 and highest in the later blocks, indicating that later blocks need fewer denoising steps once earlier context has been committed.
We use this observation by decreasing the end threshold across blocks.

\snippetbox{Block-Specific End Thresholds}{
\noindent
Let $b \in \{1,2,3,4\}$ be the block index,
$\theta_s = 0.90$ the shared start threshold, and
$w = 0.15$ the warmup fraction.
The end threshold for block $b$ is
\[
    \theta_{\mathrm{e}}^{(b)}
    = 0.70 - 0.10\,(b-1),
\]
giving $\theta_{\mathrm{e}}^{(b)} \in
\{0.70,\, 0.60,\, 0.50,\, 0.40\}$
for blocks $b=1,2,3,4$ respectively.
The commit threshold at step $t$ is
\[
\theta^{(b)}(t) =
\begin{cases}
    \theta_s
    & t < wT, \\[6pt]
    \theta_{\mathrm{e}}^{(b)} + \Delta^{(b)}(t)
    & t \geq wT,
\end{cases}
\]
where
\[
\Delta^{(b)}(t)
=
\frac{\theta_s - \theta_{\mathrm{e}}^{(b)}}{2}
\left(1 + \cos\frac{\pi(t-wT)}{T-wT}\right)
\]
and $T$ is the total steps per block.
Block~1 keeps the highest end threshold as it has the
least decoded context. Each later block lowers it
because later blocks converge more easily
(Figure~\ref{fig:bg_motivation_combined}(b)).
}{snip:block_schedule}

\subsection{Token Budgets and Position Aware Thresholds}
\label{sec:token_budget}
Adaptive thresholds reduce many unnecessary steps, but late 
denoising can still become inefficient. In this phase, the 
model may run several forward passes while committing very 
few new tokens. We call this the grinding phase. To reduce 
this cost, \CAI{} assigns each token a maximum denoising 
budget. The budget depends on the first-step confidence and 
the token position inside the block. Per token compute budgeting draws on the adaptive computation
literature \citep{graves2016act, schuster2022calm}; our key
distinction is that \textsc{CAI-dLLM} uses the first-step confidence
$c_i^{(0)}$ as a \emph{free} control signal,  no additional
forward passes are needed beyond the one required by standard
decoding.

\snippetbox{Confidence-Gated Token Budget}{
Let $s_i=c_i^{(0)}$ be the first-step confidence of token $i$.
Let $p=\mathrm{pos}(i)$ be the position of token $i$ inside its block.
The token budget $B_i$ is the maximum number of denoising steps allowed for token $i$.
Cases are evaluated from top to bottom.

\begin{equation}
B_i =
\begin{cases}
8, & s_i > 0.50 \ \text{and}\ p < 32,\\
64, & s_i < 0.25 \ \text{or}\ p \geq 48,\\
32, & \text{otherwise}.
\end{cases}
\label{eq:token_budget}
\end{equation}

If token $i$ is still masked after $B_i$ steps, it is committed using its current best prediction:
\[
x_i \leftarrow \arg\max_{v\in\mathcal{V}} p_\theta(x_i=v\mid x^{(t)}).
\]
}{snip:budget}

The budget levels follow the stability tiers in Figure~\ref{fig:bg_motivation_combined}(c). Our third observation is that token position also matters : tokens at $p < 32$ commit 37\% faster on average than those at $p \geq 48$ on GSM8K with LLaDA-8B, motivating the position-aware threshold in Box~\ref{snip:position_rule}.

\snippetbox{Position-Aware Commit Rule}{
Let $p=\mathrm{pos}(i)$ be the position of token $i$ inside its block.
Let $\theta_b(t)$ be the block-level threshold at denoising step $t$.
The token-specific threshold is
\begin{equation}
\theta_i(t)=
\mathrm{clip}\left(
\theta_b(t)\cdot r_{\mathrm{pos}(i)}, 0.20, 0.98
\right).
\label{eq:position_threshold}
\end{equation}
Here, $r_{\mathrm{pos}(i)}$ is a position factor:
\[
r_{\mathrm{pos}(i)} =
\begin{cases}
0.80, & p < 16,\\
0.90, & 16 \leq p < 32,\\
1.05, & 32 \leq p < 48,\\
1.10, & p \geq 48.
\end{cases}
\]

The final token commit rule is
\begin{equation}
c_i^{(t)} \geq \theta_i(t)
\quad \text{or} \quad
t \geq B_i,
\label{eq:position_commit}
\end{equation}
where $B_i$ is the token budget from Box~\ref{snip:budget}.
}{snip:position_rule}

The position factors are selected from a 200 sample held out subset of LLaDA GSM8K to improve speed while keeping accuracy degradation below one point.
We use the same factors for Dream-7B without further tuning. More details are provided in Appendix~\ref{app:hyperparam_justification}.

\subsection{Grinding Phase Detection}
\label{sec:grinding}

Token budgets act at the token level.
We also use a block level detector to stop low yield denoising.
At each step, we count how many tokens were committed.
If this count remains low for several consecutive steps, the block is treated as being in the grinding phase.

\snippetbox{Grinding Phase Detector}{
Let $\eta^{(t)}$ be the number of tokens committed at step $t$.
A block enters the grinding phase when

\begin{equation}
  \eta^{(t)} < \bar{\eta} \quad \text{for } K \text{ consecutive steps}
  \label{eq:grinding-detector}
\end{equation}
where $\eta^{(t)} = |\{i : x_i^{(t)} \neq \texttt{[MASK]},\,
x_i^{(t-1)} = \texttt{[MASK]}\}| \,/\, L$ is the fraction of
newly committed tokens at step $t$, $\bar{\eta}$ is the low yield
threshold (default $\bar{\eta} = 1.5\%$), $K = 4$ is the patience
window, and $L$ is the sequence length.

We use

\[
\bar{\eta}=1.5,
\qquad
K=4.
\]

When the detector triggers, remaining masked tokens are committed only if

\[
c_i^{(t)} > \theta_{\mathrm{fb}},
\qquad
\theta_{\mathrm{fb}}=0.40.
\]

The threshold $\bar{\eta}=1.5$ separates normal progress from low yield denoising.
The window $K=4$ avoids reacting to one noisy step.
The fallback threshold keeps very uncertain tokens from being forced.
}{snip:grinding}

\subsection{Model Adaptive Confidence Gating}
\label{sec:model_gating}
Different models respond differently to confidence gating.
When gating is ON, the token budgets $B_i$ from Box~\ref{snip:budget} and the force commit threshold $\theta_{fb}$ from Box~\ref{snip:grinding} are active.
When gating is OFF, only the adaptive threshold and block schedules are used. Confidence gating is set via GSM8K ablation (Appendix~\ref{app:gsm8k_ablation}): ON for LLaDA, OFF for Dream, consistent with the HumanEval ablation in Section~\ref{sec:ablation}.

\subsection{Algorithm}
\label{sec:algorithm}
Algorithm~\ref{alg:cai} summarizes \textsc{CAI-dLLM} 
for one block. Line~1 initializes the low efficiency 
counter $\ell$. Lines~3--8 run the step-zero forward 
pass to store confidence scores $s_i$, assign budgets 
$B_i$, and set the block threshold $\theta_e^{(b)}$. 
Lines~10--18 commit each masked token when its 
confidence meets the position aware threshold or its 
budget is exhausted. Lines~19--23 increment or reset 
$\ell$ based on the newly committed count $\eta^{(t)}$. 
When $\ell \geq K$ (lines~24--29), remaining 
high confidence tokens are force committed and the 
block exits early. The final stopping condition exits if all tokens are committed before the budget runs out.

\begin{algorithm}[h]
\small
\DontPrintSemicolon
\SetAlgoLined
\LinesNumbered
\SetAlgoNlRelativeSize{-1}
\SetKwInOut{Input}{Input}
\SetKwInOut{Output}{Output}
\caption{\textsc{CAI-dLLM} inference for one block}
\label{alg:cai}
\Input{$\mathbf{x}^{(0)}$: masked sequence;\ 
       $b$: block index;\ 
       $T$: max steps per block;\ 
       $\theta_s, \theta_e^{(b)}$: start/end thresholds;\
       $\theta_{fb}$: force commit threshold;\
       $w$: warmup fraction;\
       $K$: patience window;\
       $\bar{\eta}$: yield threshold}
\Output{Decoded block}
\tcp{$t \in \{0,\ldots,T-1\}$ is local to this block; $B_i$ in local steps}
$\ell \leftarrow 0$
\tcp*{$\ell$: low-efficiency step counter}
\For{$t = 0$ \KwTo $T-1$}{
    Run forward pass on $\mathbf{x}^{(t)}$\;
    Compute $c_i^{(t)}$ for all masked tokens $i$\;
    \If(\tcp*[f]{step-zero init}){$t = 0$}{
        $s_i \leftarrow c_i^{(0)}$
        \tcp*{store first-step confidence}
        Assign budget $B_i$ per token 
            (Box~\ref{snip:budget})\;
        Set threshold $\theta_e^{(b)}$ 
            (Box~\ref{snip:block_schedule})\;
    }
    Compute $\theta^{(b)}(t)$
        (Box~\ref{snip:apd});\ \
    $\eta^{(t)} \leftarrow 0$
    \tcp*{newly committed count}
    \ForEach{masked token $i$}{
        Compute $\theta_i(t)$
            (Box~\ref{snip:position_rule})\;
        \If{$c_i^{(t)} \geq \theta_i(t)$ \textbf{or} 
            $t \geq B_i$}{
            $x_i \leftarrow \arg\max_{v \in \mathcal{V}}
                \, p_\theta(x_i {=} v \mid 
                \mathbf{x}^{(t)})$\;
            $\eta^{(t)} \mathrel{+}= 1$\;
        }
    }
    \eIf(\tcp*[f]{track low-yield steps})
        {$\eta^{(t)} < \bar{\eta}$}
        {$\ell \mathrel{+}= 1$\;}
        {$\ell \leftarrow 0$\;}
    \If(\tcp*[f]{grinding phase detected})
        {$\ell \geq K$}{
        \ForEach{masked $i$ with 
            $c_i^{(t)} > \theta_{fb}$}{
            $x_i \leftarrow \arg\max_{v \in \mathcal{V}}
                \, p_\theta(x_i {=} v \mid 
                \mathbf{x}^{(t)})$\;
        }
        \textbf{break}\;
    }
    \lIf{no masked tokens remain}{\textbf{break}}
}
\Return decoded block\;
\end{algorithm}
\section{Implementation Details}
\label{sec:implementation}

\paragraph{Setup.}
\textsc{CAI-dLLM} is developed on top of the open-source PyTorch based ES-dLLM codebase, reusing its KV cache backend and integrating our confidence aware controller. We evaluate LLaDA-8B-Instruct and Dream-7B-Instruct on a single NVIDIA H200 SXM5 GPU with 141 GB HBM3e~\citep{nvidia2024h200}.
We generate 256 tokens for reasoning and commonsense tasks, and 512 tokens for code and long-context tasks.
The block length is 64 for all tasks except LongBench, where we use 32. All runs use batch size 8 and temperature 0.


\paragraph{Decoding settings.}
No-cache decoding uses 64 denoising steps per block without KV caching or parallel decoding.
All methods use the same generation settings for fair comparison.
For \textsc{CAI-dLLM}, we set the warmup fraction to $w=0.15$, the start threshold to $\theta_s=0.90$, and the base end threshold to $\theta_e=0.70$.
Confidence gating is enabled for LLaDA and disabled for Dream based on GSM8K ablations.
Following ES-dLLM, the KV cache update frequency is 16 for LLaDA and 8 for Dream~\citep{zhu2026esdllm}.

\paragraph{Baselines and metrics.}
We compare with no-cache decoding, DualCache from Fast-dLLM~\citep{wu2025fastdllm}, and ES-dLLM~\citep{zhu2026esdllm} baselines under the same task settings. For evaluation, we use exact match for GSM8K, MathQA, and BBH, pass@1 for HumanEval and MBPP, and F1/ROUGE-L for LongBench.
Throughput is output tokens per second after one warm-up pass, and speedup is wall-clock time relative to no-cache decoding. Energy is computed as $E=\bar{P}\times T$, where $\bar{P}$ is average GPU power from \texttt{nvidia-smi} and $T$ is wall-clock generation time.
Full settings, sample sizes, baselines, and software versions are in Appendix~\ref{app:setup}.

\section{Results}
\label{sec:results}
\subsection{Main Results}
\label{sec:analysis}

We evaluate on seven benchmark datasets covering math, code, reasoning, commonsense, and long-context generation~\citep{chen2021humaneval,suzgun2022bbh,austin2021mbpp,amini2019mathqa,bisk2020piqa,sakaguchi2020winogrande,bai2023longbench}.
Tables~\ref{tab:dream} and~\ref{tab:llada} report the main results.

\begin{figure*}[ht]
  \centering
  \includegraphics[width=0.95\textwidth]{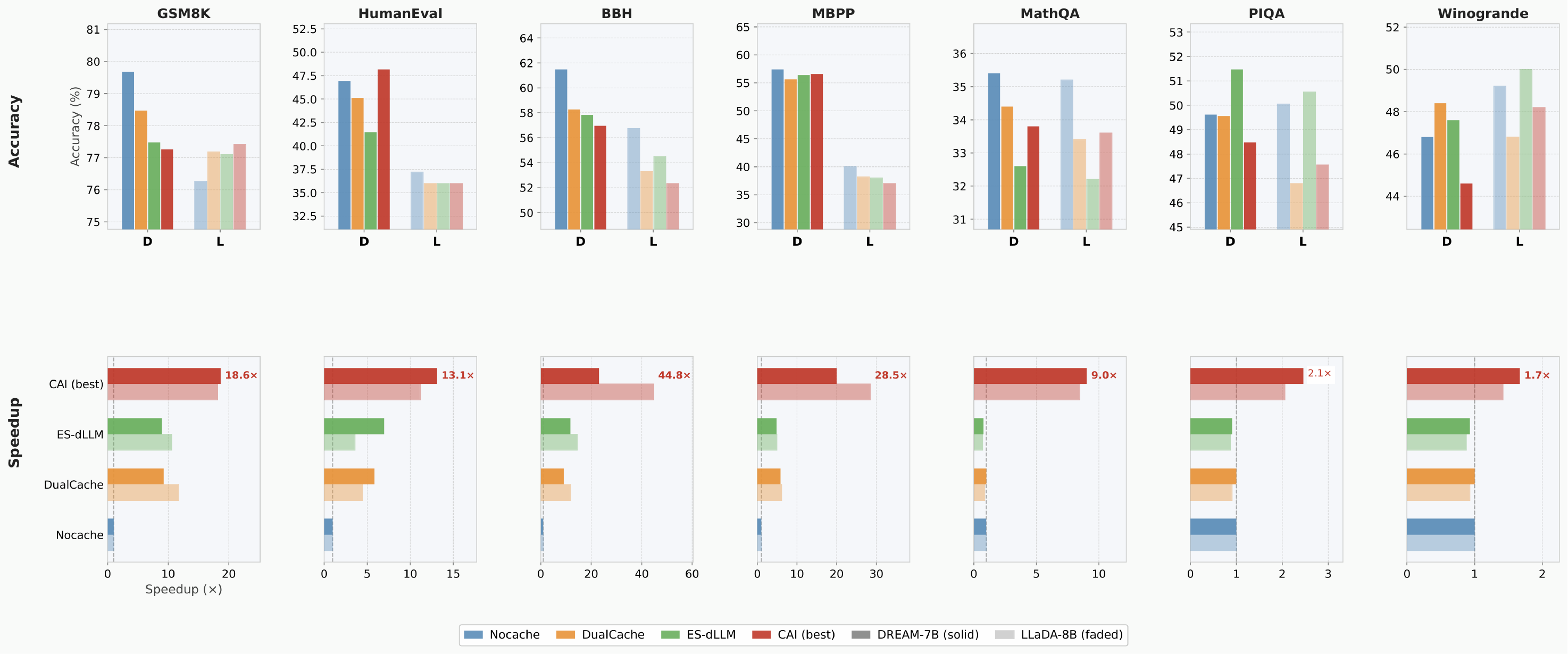}
  \caption{
  Accuracy and speedup comparison across the seven main benchmarks for Dream-7B and LLaDA-8B.
  LongBench is reported in detail in Appendix~\ref{app:longbench_results}.
  }
  \label{fig:main_accuracy_speedup}
\end{figure*}

\begin{table}[ht]
\centering
\setlength{\tabcolsep}{2.5pt}
\footnotesize
\begin{tabular}{@{}llrrrr@{}}
\toprule
\textbf{Bench.} & \textbf{Method} & \textbf{TPS}
  & \textbf{Spd.} & \textbf{Score} & \textbf{Eng.} \\
\midrule
\multirow{4}{*}{GSM8K}
 & No Cache  & 19.8  & $1.0\times$  & 79.68 & 3052 \\
 & DualCache & 103.2 & $9.3\times$  & 78.47 & 291 \\
 & ES-dLLM   & 97.9  & $9.0\times$  & 77.48 & 219 \\
\rowcolor{caihighlight}
 & \textbf{\textsc{CAI-dLLM}} & \textbf{369.5} & $\mathbf{18.7\times}$
   & $77.26{\pm}1.18$ & \textbf{146} \\
\midrule
\multirow{4}{*}{\shortstack[l]{Human\\Eval}}
 & No Cache  & 44.3  & $1.0\times$  & 46.95 & 374 \\
 & DualCache & 258.1 & $5.8\times$  & 45.12 & 60 \\
 & ES-dLLM   & 308.5 & $7.0\times$  & 41.46 & 41 \\
\rowcolor{caihighlight}
 & \textbf{\textsc{CAI-dLLM}} & \textbf{581.5} & $\mathbf{13.1\times}$
   & $48.17{\pm}3.76$ & \textbf{24} \\
\midrule
\multirow{4}{*}{BBH}
 & No Cache  & 24.8  & $1.0\times$  & 61.48 & 12655 \\
 & DualCache & 226.9 & $9.1\times$  & 58.27 & 1318 \\
 & ES-dLLM   & 292.2 & $11.8\times$ & 57.84 & 927 \\
\rowcolor{caihighlight}
 & \textbf{\textsc{CAI-dLLM}} & \textbf{571.4} & $\mathbf{23.0\times}$
   & $56.95{\pm}0.53$ & \textbf{500} \\
\midrule
\multirow{4}{*}{MBPP}
 & No Cache  & 29.1  & $1.0\times$  & 57.40 & 868 \\
 & DualCache & 170.4 & $5.9\times$  & 55.40 & 104 \\
 & ES-dLLM   & 141.0 & $4.9\times$  & 56.40 & 78 \\
\rowcolor{caihighlight}
 & \textbf{\textsc{CAI-dLLM}} & \textbf{581.7} & $\mathbf{20.0\times}$
   & $56.60{\pm}2.16$ & \textbf{34} \\
\midrule
\multirow{4}{*}{MathQA}
 & No Cache  & 58.8  & $1.0\times$  & 35.40 & 54 \\
 & DualCache & 59.1  & $1.0\times$  & 34.40 & 43 \\
 & ES-dLLM   & 45.2  & $0.8\times$  & 32.60 & 47 \\
\rowcolor{caihighlight}
 & \textbf{\textsc{CAI-dLLM}} & \textbf{530.8} & $\mathbf{9.0\times}$
   & $33.80{\pm}2.13$ & \textbf{9} \\
\midrule
\multirow{4}{*}{PIQA}
 & No Cache  & 62.5  & $1.0\times$  & 49.62 & 158 \\
 & DualCache & 62.7  & $1.0\times$  & 49.56 & 140 \\
 & ES-dLLM   & 57.1  & $0.9\times$  & 51.47 & 135 \\
\rowcolor{caihighlight}
 & \textbf{\textsc{CAI-dLLM}} & \textbf{130.9} & $\mathbf{2.1\times}$
   & $49.08{\pm}1.17$ & \textbf{61} \\
\midrule
\multirow{4}{*}{\shortstack[l]{Wino-\\grande}}
 & No Cache  & 64.6  & $1.0\times$  & 46.80 & 42 \\
 & DualCache & 64.7  & $1.0\times$  & 48.40 & 39 \\
 & ES-dLLM   & 59.8  & $0.9\times$  & 47.60 & 38 \\
\rowcolor{caihighlight}
 & \textbf{\textsc{CAI-dLLM}} & \textbf{107.6} & $\mathbf{1.7\times}$
   & $44.60{\pm}2.22$ & \textbf{24} \\
\bottomrule
\end{tabular}
\caption{Results on Dream-7B-Instruct. TPS = tokens/sec;
Eng.\ = energy in Wh.
\textbf{Bold} = best. DualCache from \citet{wu2025fastdllm}.
\textsc{CAI-dLLM} scores show std.\ err.\ as $\pm$.}
\label{tab:dream}
\end{table}

\begin{table}[t]
\centering
\setlength{\tabcolsep}{2.5pt}
\footnotesize
\begin{tabular}{@{}llrrrr@{}}
\toprule
\textbf{Bench.} & \textbf{Method} & \textbf{TPS}
  & \textbf{Spd.} & \textbf{Score} & \textbf{Eng.} \\
\midrule
\multirow{4}{*}{GSM8K}
 & No Cache  & 17.1  & $1.0\times$  & 76.27 & 3559 \\
 & DualCache & 196.4 & $11.7\times$ & \textbf{77.18} & 264 \\
 & ES-dLLM   & 175.3 & $10.6\times$ & 77.10 & 223 \\
\rowcolor{caihighlight}
 & \textbf{\textsc{CAI-dLLM}} & \textbf{335.1} & $\mathbf{18.2\times}$
   & $77.41{\pm}1.18$ & \textbf{169} \\
\midrule
\multirow{4}{*}{\shortstack[l]{Human\\Eval}}
 & No Cache  & 39.0  & $1.0\times$  & 37.20 & 425 \\
 & DualCache & 173.0 & $4.4\times$  & 35.98 & 55 \\
 & ES-dLLM   & 139.8 & $3.6\times$  & 35.98 & 42 \\
\rowcolor{caihighlight}
 & \textbf{\textsc{CAI-dLLM}} & \textbf{435.3} & $\mathbf{11.2\times}$
   & $35.98{\pm}3.76$ & \textbf{33} \\
\midrule
\multirow{4}{*}{BBH}
 & No Cache  & 11.1  & $1.0\times$  & 56.75 & 28425 \\
 & DualCache & 130.1 & $11.8\times$ & 53.29 & 1193 \\
 & ES-dLLM   & 168.8 & $14.5\times$ & 54.51 & 915 \\
\rowcolor{caihighlight}
 & \textbf{\textsc{CAI-dLLM}} & \textbf{495.1} & $\mathbf{44.8\times}$
   & $52.33{\pm}0.53$ & \textbf{579} \\
\midrule
\multirow{4}{*}{MBPP}
 & No Cache  & 15.3  & $1.0\times$  & 40.00 & 1029 \\
 & DualCache & 93.7  & $6.1\times$  & 38.20 & 97 \\
 & ES-dLLM   & 75.6  & $4.9\times$  & 38.00 & 81 \\
\rowcolor{caihighlight}
 & \textbf{\textsc{CAI-dLLM}} & \textbf{435.6} & $\mathbf{28.5\times}$
   & $37.00{\pm}2.16$ & \textbf{51} \\
\midrule
\multirow{4}{*}{MathQA}
 & No Cache  & 48.5  & $1.0\times$  & 35.20 & 68 \\
 & DualCache & 42.5  & $0.9\times$  & 33.40 & 49 \\
 & ES-dLLM   & 34.1  & $0.7\times$  & 32.20 & 56 \\
\rowcolor{caihighlight}
 & \textbf{\textsc{CAI-dLLM}} & \textbf{411.4} & $\mathbf{8.5\times}$
   & $33.60{\pm}2.13$ & \textbf{10} \\
\midrule
\multirow{4}{*}{PIQA}
 & No Cache  & 51.0  & $1.0\times$  & 50.05 & 178 \\
 & DualCache & 46.6  & $0.9\times$  & 46.79 & 163 \\
 & ES-dLLM   & 44.5  & $0.9\times$  & 50.54 & 162 \\
\rowcolor{caihighlight}
 & \textbf{\textsc{CAI-dLLM}} & \textbf{105.3} & $\mathbf{2.1\times}$
   & $47.55{\pm}1.17$ & \textbf{82} \\
\midrule
\multirow{4}{*}{\shortstack[l]{Wino-\\grande}}
 & No Cache  & 52.4  & $1.0\times$  & 49.20 & 48 \\
 & DualCache & 48.5  & $0.9\times$  & 46.80 & 45 \\
 & ES-dLLM   & 46.2  & $0.9\times$  & 50.00 & 44 \\
\rowcolor{caihighlight}
 & \textbf{\textsc{CAI-dLLM}} & \textbf{74.3} & $\mathbf{1.4\times}$
   & $48.20{\pm}2.23$ & \textbf{31} \\
\bottomrule
\end{tabular}
\caption{Results on LLaDA-8B-Instruct. TPS = tokens/sec;
Eng.\ = energy in Wh.
\textbf{Bold} = best. Sample sizes in Appendix~\ref{app:setup}.
\textsc{CAI-dLLM} scores show std.\ err.\ as $\pm$.}
\label{tab:llada}
\end{table}

\textsc{CAI-dLLM} achieves the highest throughput on every evaluated dataset (Figure~\ref{fig:main_accuracy_speedup}). Gains scale with generation length: on longer tasks, speedup reaches $44.8\times$ on LLaDA BBH, $28.5\times$ on LLaDA MBPP, $23.0\times$ on Dream BBH, $20.0\times$ on Dream MBPP, and $19.0\times$ on Dream LongBench. This length-scaling trend holds across generation lengths, with $4.59\times$ on LLaDA-Instruct and $15.29\times$ on Dream-Instruct at length 512 (Appendix~\ref{app:throughput}). On shorter commonsense tasks (PIQA, Winogrande), \textsc{CAI-dLLM} remains the only method with clear speedup, though at a modest accuracy trade-off versus ES-dLLM (Appendix~\ref{app:short_commonsense}). The controller overhead is below 2\%, while early commitment and grinding-phase exit reduce per-iteration time by 13.6\% on LLaDA and 22.5\% on Dream.

Accuracy shows a clear efficiency-quality trade-off.
On Dream HumanEval, \textsc{CAI-dLLM} improves pass@1 from $46.95$ to $48.17$ while giving a $13.1\times$ speedup; with a standard error of $\pm 3.76$, we treat this as accuracy parity.
On harder reasoning tasks, the trade-off is larger.
For example, LLaDA BBH drops from $56.75$ to $52.33$ while reaching $44.8\times$ speedup.
Dream GSM8K also drops from $79.68$ to $77.26$, but gives an $18.7\times$ speedup.

\paragraph{Long-context results.}
On LongBench, evaluated with Dream-7B, \textsc{CAI-dLLM} reaches $19.04\times$ speedup with an average score of 23.11, compared with 24.63 for no-cache decoding.
All cached methods slightly reduce the average score in this setting, but \textsc{CAI-dLLM} provides more than twice their throughput.
Energy follows the same trend as runtime.
For example, on LLaDA GSM8K, \textsc{CAI-dLLM} reduces energy from $3559$ Wh to $169$ Wh, a $95.3\%$ reduction.

\subsection{Ablation Study}
\label{sec:ablation}

We ablate the main components of \textsc{CAI-dLLM} on HumanEval.
Table~\ref{tab:humaneval_ablation} reports results for both models.
GSM8K model adaptation results are reported in Appendix~\ref{app:gsm8k_ablation}, Table~\ref{tab:gsm8k_ablation}.

\paragraph{LLaDA-8B-Instruct.}
APD gives the largest throughput gain, increasing throughput from 139.8 TPS with ES-dLLM to 424.3 TPS.
This comes with a small pass@1 drop from 35.98\% to 35.37\%.
Adding the per-block schedule gives a small further speedup, reaching 434.2 TPS with the same pass@1.
Adding token budgets gives the highest throughput for LLaDA, reaching 435.3 TPS, while keeping pass@1 at 35.37\%.
This shows that most of the gain comes from APD, while block scheduling and budgets add smaller improvements.

\paragraph{Dream-7B-Instruct.}
For Dream, the gated budget configuration reaches very high throughput, but it lowers pass@1 to 40.34\%.
Disabling confidence gating recovers pass@1 to 46.34\%, which is higher than DualCache and close to the no-cache baseline, while keeping 614.6 TPS.
This supports our model-adaptive policy: LLaDA can use confidence-gated budgets, while Dream performs better when gating is disabled.

\begin{table}[!t]
  \centering
  \footnotesize
  \setlength{\tabcolsep}{3.5pt}
  \renewcommand{\arraystretch}{0.92}
  \begin{tabular}{llcc}
    \hline
    \textbf{Model} & \textbf{Config.} & \textbf{Pass@1} & \textbf{TPS} \\
    \hline
    \multirow{4}{*}{LLaDA}
      & ES-dLLM                  & 35.98 & 139.8 \\
      & +APD                     & 35.37 & 424.3 \\
      & +APD+Block               & 35.37 & 434.2 \\
      & +APD+Block+Budget $\star$& 35.37 & 435.3 \\
    \hline
    \multirow{4}{*}{Dream}
      & DualCache                & 45.12 & 72.5  \\
      & ES-dLLM                  & 41.46 & 56.7  \\
      & +APD+Block+Budget        & 40.34 & 614.7 \\
      & Full CAI, gate off $\star$ & 46.34 & 614.6 \\
    \hline
  \end{tabular}
  \caption{HumanEval ablation of \textsc{CAI-dLLM} components. $\star$ marks the recommended configuration for each model.}
  \label{tab:humaneval_ablation}
\end{table}

\paragraph{Denoising-step reduction.}
Table~\ref{tab:step_reduction} confirms that \textsc{CAI-dLLM} reduces denoising iterations directly, not only per-step cost, with 39--56\% step reduction on GSM8K and 38--47\% on HumanEval.

\begin{table}[!t]
  \centering
  \small
  \setlength{\tabcolsep}{4pt}
  \renewcommand{\arraystretch}{0.95}
  \begin{tabular}{llccc}
    \hline
    \textbf{Task} & \textbf{Model} & \textbf{Fixed} & \textbf{CAI} & \textbf{Reduction} \\
    \hline
    GSM8K     & LLaDA & 256 & 114 & 55.5\% \\
    GSM8K     & Dream & 256 & 141 & 44.9\% \\
    HumanEval & LLaDA & 512 & 315 & 38.5\% \\
    HumanEval & Dream & 512 & 271 & 47.1\% \\
    \hline
  \end{tabular}
  \caption{Average denoising steps per request and step reduction relative to fixed budget.}
  \label{tab:step_reduction}
\end{table}

\section{Conclusion}
We present \textsc{CAI-dLLM}, a  training-free inference method 
for diffusion language models that uses first-step confidence to 
guide token commitment, block schedules, and token budgets without 
extra models, retraining, or weight updates. Across seven benchmarks 
on LLaDA-8B-Instruct and Dream-7B-Instruct, \textsc{CAI-dLLM} 
achieves consistent speedups, up to 44.8$\times$, with the 
largest gains on longer generation tasks and up to 95.3\% energy 
reduction. These results show that first-step confidence, available 
at no extra cost, is sufficient to substantially reduce redundant 
denoising computation while keeping the original model unchanged.
\section*{Limitations}

\textsc{CAI-dLLM} has several limitations. First, our model-specific confidence gating choice is selected using GSM8K ablations on the full test set: gating is enabled for LLaDA-8B and disabled for Dream-7B. This is a limitation of the current study because the gating decision is not selected on a held out validation set. For deployment on a new model, the gating choice should be selected on a small held out validation subset of a representative task, such as 200--500 samples. Second, the speedups can come with accuracy trade-offs. For example, accuracy drops from 56.75 to 52.33 on BBH with LLaDA-8B, and from 40.00 to 37.00 on MBPP with LLaDA-8B. This makes the method more suitable for settings where lower latency or energy use is important and moderate quality loss is acceptable. Third, all experiments are run on a single NVIDIA H200 GPU, so results may differ on other hardware. Finally, some evaluations use reduced sample sizes to control computation cost, such as 500 samples for MathQA and reduced task subsets for LongBench, which may increase variance.



\section*{Ethical Considerations}

This work presents a training-free inference acceleration method for masked diffusion language models. It does not add new model capabilities, change model weights, or expand the types of content the models can generate. The main goal is to reduce inference cost and energy use for existing diffusion language model deployments. The ethical risks of this work are therefore mainly the risks of the underlying models, such as LLaDA-8B and Dream-7B. Standard responsible-use practices for large language models still apply. Generative AI tools were used only for language polishing and editing of author written text. They were not used to generate experimental results, technical claims, citations, or research conclusions. All authors reviewed and approved the final manuscript and are responsible for its content. All technical claims, experiments, citations, and final text were checked and approved by the authors.

\paragraph{Code Release.}
We will release the complete \textsc{CAI-dLLM} implementation, 
including inference scripts, configuration files, and 
evaluation code, upon acceptance of this paper.
\bibliography{custom}

@article{grattafiori2024llama,
  title   = {The Llama 3 Herd of Models},
  author  = {Grattafiori, Aaron and Dubey, Abhimanyu and Jauhri, Abhinav and Pandey, Abhinav and Kadian, Abhishek and Al-Dahle, Ahmad and others},
  journal = {arXiv preprint arXiv:2407.21783},
  year    = {2024}
}

@article{liu2025dllmcache,
  title   = {dLLM Cache: Accelerating Diffusion Large Language Models with Adaptive Caching},
  author  = {Liu, Zhiyuan and Yang, Yicun and Zhang, Yaojie and Chen, Junjie and Zou, Chang and Wei, Qingyan and Wang, Shaobo and Zhang, Linfeng},
  journal = {arXiv preprint arXiv:2506.06295},
  year    = {2025}
}

@article{song2025sparsedllm,
  title   = {Sparse dLLM: Accelerating Diffusion LLMs with Dynamic Cache Eviction},
  author  = {Song, Yuerong and Liu, Xiaoran and Li, Ruixiao and Liu, Zhigeng and Huang, Zengfeng and Guo, Qipeng and He, Ziwei and Qiu, Xipeng},
  journal = {arXiv preprint arXiv:2508.02558},
  year    = {2025}
}

@article{ye2025dream,
  title   = {{Dream 7B}: Diffusion Large Language Models},
  author  = {Ye, Jiacheng and Xie, Zhihui and Zheng, Lin and 
             Gao, Jiahui and Wu, Zirui and Jiang, Xin and 
             Li, Zhenguo and Kong, Lingpeng},
  journal = {arXiv preprint arXiv:2508.15487},
  year    = {2025}
}

@inproceedings{cobbe2021gsm8k,
  title     = {Training Verifiers to Solve Math Word Problems},
  author    = {Cobbe, Karl and Kosaraju, Vineet and Bavarian, Mohammad and Chen, Mark and Jun, Heewoo and Kaiser, Lukasz and Plappert, Matthias and Tworek, Jerry and Hilton, Jacob and Nakano, Reiichiro and Hesse, Christopher and Schulman, John},
  booktitle = {arXiv preprint arXiv:2110.14168},
  year      = {2021}
}

@inproceedings{zhu2026esdllm,
  title     = {ES-DLLM: Efficient Inference for Diffusion Large Language Models by Early-Skipping},
  author    = {Zhu, Zijian and Ren, Fei and Tan, Zhanhong and Ma, Kaisheng},
  booktitle = {International Conference on Learning Representations},
  year      = {2026}
}

@article{nie2025llada,
  title   = {Large Language Diffusion Models},
  author  = {Nie, Shen and Zhu, Fengqi and You, Zebin and Zhang, Xiaolu and Ou, Jingyang and Hu, Jun and Zhou, Jun and Lin, Yankai and Wen, Ji Rong and Li, Chongxuan},
  journal = {arXiv preprint arXiv:2502.09992},
  year    = {2025}
}

@article{sahoo2024simple,
  title   = {Simple and Effective Masked Diffusion Language Models},
  author  = {Sahoo, Subham Sekhar and Arriola, Mariano and Schiff, Yair and Gokaslan, Aaron and Marroquin, Edgar and Chiu, Justin T. and Kuleshov, Volodymyr},
  journal = {arXiv preprint arXiv:2406.07524},
  year    = {2024}
}

@article{ma2025dkvcache,
  title   = {dKV Cache: The Cache for Diffusion Language Models},
  author  = {Ma, Xinyin and Yu, Runpeng and Fang, Gongfan and Wang, Xinchao},
  journal = {arXiv preprint arXiv:2505.15781},
  year    = {2025}
}

@article{hu2025flashdlm,
  title   = {FlashDLM: Accelerating Diffusion Language Model Inference via Efficient KV Caching and Guided Diffusion},
  author  = {Hu, Zhanqiu and Meng, Jian and Akhauri, Yash and Abdelfattah, Mohamed S. and Seo, Jae Sun and Zhang, Zhiru and Gupta, Udit},
  journal = {arXiv preprint arXiv:2505.21467},
  year    = {2025}
}

@article{agrawal2026spiffy,
  title   = {Spiffy: Multiplying Diffusion LLM Acceleration via Lossless Speculative Decoding},
  author  = {Agrawal, Sudhanshu and Garrepalli, Risheek and Goel, Raghavv and Lee, Mingu and Lott, Christopher and Porikli, Fatih},
  journal = {arXiv preprint arXiv:2509.18085},
  year    = {2025}
}

@inproceedings{vaswani2017attention,
  title     = {Attention Is All You Need},
  author    = {Vaswani, Ashish and Shazeer, Noam and Parmar, Niki and Uszkoreit, Jakob and Jones, Llion and Gomez, Aidan N. and Kaiser, Lukasz and Polosukhin, Illia},
  booktitle = {Advances in Neural Information Processing Systems},
  year      = {2017}
}

@inproceedings{brown2020language,
  title     = {Language Models are Few Shot Learners},
  author    = {Brown, Tom B. and Mann, Benjamin and Ryder, Nick and Subbiah, Melanie and Kaplan, Jared and Dhariwal, Prafulla and Neelakantan, Arvind and Shyam, Pranav and Sastry, Girish and Askell, Amanda and others},
  booktitle = {Advances in Neural Information Processing Systems},
  year      = {2020}
}

@inproceedings{austin2021structured,
  title     = {Structured Denoising Diffusion Models in Discrete State Spaces},
  author    = {Austin, Jacob and Johnson, Daniel D. and Ho, Jonathan and Tarlow, Daniel and van den Berg, Rianne},
  booktitle = {Advances in Neural Information Processing Systems},
  year      = {2021}
}

@inproceedings{lou2023discrete,
  title     = {Discrete Diffusion Modeling by Estimating the Ratios of the Data Distribution},
  author    = {Lou, Aaron and Meng, Chenlin and Ermon, Stefano},
  booktitle = {International Conference on Machine Learning},
  year      = {2023}
}

@article{wu2025fastdllm,
  title   = {Fast dLLM: Training Free Acceleration of Diffusion LLM by Enabling KV Cache and Parallel Decoding},
  author  = {Wu, Chengyue and Zhang, Hao and Xue, Shuchen and Liu, Zhijian and Diao, Shizhe and Zhu, Ligeng and Luo, Ping and Han, Song and Xie, Enze},
  journal = {arXiv preprint arXiv:2505.22618},
  year    = {2025}
}

@article{wang2025d2f,
  title   = {Diffusion LLMs Can Do Faster than AR Inference via Discrete Diffusion Forcing},
  author  = {Wang, Xu and Xu, Chenkai and Jin, Yijie and Jin, Jiachun and Zhang, Hao and Deng, Zhijie},
  journal = {arXiv preprint arXiv:2508.09192},
  year    = {2025}
}

@inproceedings{chen2021humaneval,
  title     = {Evaluating Large Language Models Trained on Code},
  author    = {Chen, Mark and Tworek, Jerry and Jun, Heewoo and Yuan, Qiming
               and Pinto, Henrique Ponde de Oliveira and Kaplan, Jared and
               Edwards, Harri and Burda, Yuri and Joseph, Nicholas and
               Brockman, Greg and others},
  booktitle = {arXiv preprint arXiv:2107.03374},
  year      = {2021}
}

@inproceedings{suzgun2022bbh,
  title     = {Challenging {BIG}-Bench Tasks and Whether Chain-of-Thought
               Can Solve Them},
  author    = {Suzgun, Mirac and Scales, Nathan and Sch{\"a}rli, Nathanael
               and Gehrmann, Sebastian and Tay, Yi and Chung, Hyung Won and
               Chowdhery, Aakanksha and Le, Quoc V. and Chi, Ed H. and
               Zhou, Denny and Wei, Jason},
  booktitle = {Findings of the Association for Computational Linguistics:
               ACL 2023},
  year      = {2023}
}

@article{austin2021mbpp,
  title   = {Program Synthesis with Large Language Models},
  author  = {Austin, Jacob and Odena, Augustus and Nye, Maxwell and
             Bosma, Maarten and Michalewski, Henryk and Dohan, David and
             Jiang, Ellen and Cai, Carrie J and Terry, Michael and Le,
             Quoc V. and Sutton, Charles},
  journal = {arXiv preprint arXiv:2108.07732},
  year    = {2021}
}

@inproceedings{amini2019mathqa,
  title     = {{MathQA}: Towards Interpretable Math Word Problem Solving
               with Operation-Based Formalisms},
  author    = {Amini, Aida and Gabriel, Saadia and Lin, Peter and
               Koncel-Kedziorski, Rik and Choi, Yejin and Hajishirzi, Hannaneh},
  booktitle = {Proceedings of the 2019 Conference of the North {A}merican
               Chapter of the Association for Computational Linguistics:
               Human Language Technologies},
  year      = {2019}
}

@inproceedings{bisk2020piqa,
  title     = {{PIQA}: Reasoning about Physical Commonsense in Natural
               Language},
  author    = {Bisk, Yonatan and Zellers, Rowan and Gao, Jianfeng and
               Choi, Yejin},
  booktitle = {Proceedings of the Thirty-Fourth {AAAI} Conference on
               Artificial Intelligence},
  year      = {2020}
}

@article{sakaguchi2020winogrande,
  title   = {{W}inogrande: An Adversarial Winograd Schema Challenge at Scale},
  author  = {Sakaguchi, Keisuke and Bras, Ronan Le and Bhagavatula, Chandra
             and Choi, Yejin},
  journal = {Communications of the {ACM}},
  volume  = {64},
  number  = {9},
  year    = {2020}
}

@inproceedings{bai2023longbench,
  title     = {{LongBench}: A Bilingual, Multitask Benchmark for Long Context
               Understanding},
  author    = {Bai, Yushi and Lv, Xin and Zhang, Jiajie and Lyu, Hongchang
               and Tang, Jiayi and Huang, Zhidian and Du, Zhengxiao and
               Liu, Xiao and Zeng, Aohan and Hou, Lei and Dong, Yuxiao
               and Tang, Jie and Li, Juanzi},
  booktitle = {Proceedings of the 62nd Annual Meeting of the Association
               for Computational Linguistics},
  year      = {2024}
}

@article{graves2016act,
  title   = {Adaptive Computation Time for Recurrent Neural Networks},
  author  = {Graves, Alex},
  journal = {arXiv preprint arXiv:1603.08983},
  year    = {2016}
}

@inproceedings{schuster2022calm,
  title     = {Confident Adaptive Language Modeling},
  author    = {Schuster, Tal and Fisch, Adam and Gupta, Jai and
               Dehghani, Mostafa and Bahri, Dara and Tran, Vinh Q. and
               Tay, Yi and Metzler, Donald},
  booktitle = {Advances in Neural Information Processing Systems},
  volume    = {35},
  year      = {2022}
}

@inproceedings{kwon2023vllm,
  title     = {Efficient Memory Management for Large Language Model Serving
               with {PagedAttention}},
  author    = {Kwon, Woosuk and Li, Zhuohan and Zhuang, Siyuan and Sheng,
               Ying and Zheng, Lianmin and Yu, Cody Hao and Gonzalez,
               Joseph E. and Zhang, Hao and Stoica, Ion},
  booktitle = {Proceedings of the ACM {SIGOPS} 29th Symposium on Operating
               Systems Principles},
  year      = {2023}
}

@inproceedings{lee2018iterative,
  title     = {Deterministic Non-Autoregressive Neural Sequence Modeling
               by Iterative Refinement},
  author    = {Lee, Jason and Mansimov, Elman and Cho, Kyunghyun},
  booktitle = {Proceedings of the 2018 Conference on Empirical Methods in
               Natural Language Processing},
  pages     = {1173--1182},
  year      = {2018}
}

@misc{eval-harness,
  title  = {Language Model Evaluation Harness},
  author = {Gao, Leo and Tow, Jonathan and Abbasi, Baber 
            and Biderman, Stella and Black, Sid and 
            DiPofi, Anthony and Foster, Charles and 
            Golding, Laurence and Hsu, Jeffrey and 
            Le Noac'h, Alain and Li, Haonan and 
            McDonell, Kyle and Muennighoff, Niklas and 
            Ociepa, Chris and Phang, Jason and 
            Reynolds, Laria and Schoelkopf, Hailey and 
            Skowron, Aviya and Sutawika, Lintang and 
            Tang, Eric and Thite, Anish and 
            Wang, Ben and Wang, Kevin and Zou, Andy},
  year   = {2023},
  url    = {https://github.com/EleutherAI/lm-evaluation-harness}
}

@inproceedings{leviathan2023fast,
  title     = {Fast Inference from Transformers 
               via Speculative Decoding},
  author    = {Leviathan, Yaniv and Kalman, Matan 
               and Matias, Yossi},
  booktitle = {International Conference on Machine 
               Learning},
  year      = {2023}
}

@misc{nvidia2024h200,
  title  = {{NVIDIA H200 Tensor Core GPU}},
  author = {{NVIDIA Corporation}},
  year   = {2024},
  url    = {https://www.nvidia.com/en-us/data-center/h200/}
}

\appendix
\appendix

\section{Evaluation Setup and Reproducibility Details}
\label{app:setup}

This appendix gives the detailed evaluation settings used in Section~\ref{sec:implementation}.
All compared methods use the same generation length, block length, batch size, temperature, and evaluation split for each benchmark.

\subsection{Generation Settings}
\label{app:generation_settings}

Table~\ref{tab:gen_settings} summarizes the generation settings used across benchmark groups.
For reasoning and commonsense benchmarks (GSM8K, BBH, MathQA, PIQA, WinoGrande), we generate 256 tokens.
For code benchmarks (HumanEval, MBPP) and LongBench, we generate 512 tokens.
Block length 64 follows the standard block wise decoding setup used in recent diffusion language model inference work~\citep{wu2025fastdllm,zhu2026esdllm}.
For LongBench, we use block length 32 for finer commit control over longer generations.
All experiments use batch size 8 and temperature 0.

\begin{table}[t]
\centering
\small
\setlength{\tabcolsep}{3pt}
\resizebox{\columnwidth}{!}{
\begin{tabular}{lcccc}
\toprule
\textbf{Setting} & \textbf{Reasoning} & \textbf{Code} 
  & \textbf{Commonsense} & \textbf{LongBench} \\
\midrule
Gen.\ tokens & 256 & 512 & 256 & 512 \\
Block length & 64  & 64  & 64  & 32  \\
Batch size   & 8   & 8   & 8   & 8   \\
Temperature  & 0   & 0   & 0   & 0   \\
\bottomrule
\end{tabular}}
\caption{
Generation settings, identical across all compared methods.
}
\label{tab:gen_settings}
\end{table}

For \textsc{CAI-dLLM}, the adaptive threshold schedule 
uses $w=0.15$, $\theta_s=0.90$, and $\theta_e=0.70$. 
Confidence gating refers to whether token budgets $B_i$ 
and force commitment at $\theta_{fb}$ are active (ON) 
or disabled (OFF), while keeping the adaptive threshold 
and block schedules in both cases. We enable gating for 
LLaDA and disable it for Dream because Dream's 
confidence scores are more uniform across token 
positions, making forced token commitment 
counterproductive; disabling it gives a better 
accuracy-speed trade-off on Dream as shown in 
Table~\ref{tab:gsm8k_ablation} 
(Appendix~\ref{app:ablation_results}).
. The KV cache update 
frequency follows the ES-dLLM evaluation 
scripts~\citep{zhu2026esdllm}: 16 for LLaDA and 8 for 
Dream, meaning the cache is refreshed every $f$ 
denoising steps.

\subsection{Hyperparameter Justification}
\label{app:hyperparam_justification}

\paragraph{Position aware commit rule.}
The position aware commit rule uses four factors,
$r_{\mathrm{pos}(i)} \in \{0.80, 0.90, 1.05, 1.10\}$.
These factors reflect the within block stability pattern observed in our motivation analysis.
In LLaDA GSM8K, tokens in the first quarter of a block commit faster than tokens in the last quarter.
We therefore use smaller factors for early positions and larger factors for late positions.

The factors are selected on a 200-sample held out subset of LLaDA GSM8K.
We choose the setting that improves speed while keeping accuracy degradation below one point.
The same factors are then used for Dream-7B without additional tuning.

The clipping range $(0.20, 0.98)$ acts as a safety bound.
The upper bound prevents very high thresholds from blocking commitment.
The lower bound prevents overly aggressive commitment when thresholds are changed outside the main setting.

\subsection{Benchmark Sample Sizes}
\label{app:sample_sizes}

Table~\ref{tab:sample_sizes} reports the exact number 
of samples used for each benchmark. Full test or 
validation sets are used wherever feasible. The two 
exceptions are MathQA and LongBench, where we use 
subsets to keep total GPU time within a practical 
budget: running all four methods across both models 
on the full MathQA validation set (4,475 samples) 
or the full LongBench set would require several 
additional days of H200 compute. The subset sizes 
(500 for MathQA, 100 per task for LongBench) are 
standard in prior work on diffusion language model 
inference~\citep{zhu2026esdllm} and are large enough 
to produce stable accuracy estimates, as confirmed 
by the standard errors reported in 
Table~\ref{tab:stderr}.

The subset choice reflects evaluation cost only and 
does not indicate any limitation of \textsc{CAI-dLLM} 
on longer or harder inputs. In practice, \textsc{CAI-dLLM} 
is designed precisely for varied request types: it 
assigns easy tokens fewer steps and hard tokens more, 
so it naturally adapts to short and long outputs 
within the same batch. The LongBench results 
(Appendix~\ref{app:longbench_results}) confirm that 
\textsc{CAI-dLLM} transfers to long-context inference 
with a 19.04$\times$ speedup and minimal quality loss.

\begin{table}[t]
\centering
\small
\setlength{\tabcolsep}{3pt}
\resizebox{\columnwidth}{!}{
\begin{tabular}{llrrl}
\toprule
\textbf{Benchmark} & \textbf{Split} 
  & \textbf{Used} & \textbf{Full} & \textbf{Notes} \\
\midrule
GSM8K      & test       & 1,319 & 1,319 & Full test set \\
BBH        & test       & 6,511 & 6,511 & Full; 23 subtasks \\
HumanEval  & test       & 164   & 164   & Full test set \\
MBPP       & test       & 257   & 257   & Full sanitized set \\
WinoGrande & validation & 500   & 1,267 & Standard eval split \\
PIQA       & validation & 1,838 & 1,838 & Full validation set \\
MathQA     & validation & 500   & 4,475 & Subset; eval cost \\
LongBench  & test       & 100   & 200   & Per task; 4 tasks \\
\bottomrule
\end{tabular}}
\caption{
Sample sizes per benchmark.
Subsets are used only to control evaluation cost, not due to any limitation of \textsc{CAI-dLLM}.
}
\label{tab:sample_sizes}
\end{table}

\subsection{LongBench Task Selection}
\label{app:longbench_selection}

LongBench~\citep{bai2023longbench} is evaluated on Dream-7B only.
LLaDA-8B is excluded because its maximum context length is insufficient for most LongBench tasks.
We evaluate four tasks: NarrativeQA, MultifieldQA-en, GovReport, and QMSum.
We remove HotpotQA and Qasper because they showed high variance under the 100-sample setting.
Thus, our LongBench result should be interpreted as a selected-task LongBench evaluation rather than the full LongBench suite.
All inputs are truncated to 4,000 tokens using \texttt{--max\_input\_len 4000}.
Table~\ref{tab:longbench_tasks} lists the selected tasks, metrics, sample counts, and input length limit.

\begin{table}[t]
\centering
\small
\setlength{\tabcolsep}{3pt}
\resizebox{\columnwidth}{!}{
\begin{tabular}{llcc}
\toprule
\textbf{Task} & \textbf{Metric} 
  & \textbf{Samples} & \textbf{Max input} \\
\midrule
NarrativeQA & F1 & 100 & 4,000 \\
MultifieldQA-en & F1 & 100 & 4,000 \\
GovReport & ROUGE-L & 100 & 4,000 \\
QMSum & ROUGE-L & 100 & 4,000 \\
\midrule
Total & -- & 400 & -- \\
\bottomrule
\end{tabular}}
\caption{
LongBench tasks used in evaluation.
}
\label{tab:longbench_tasks}
\end{table}

\subsection{Evaluation Metrics}
\label{app:metrics}

Table~\ref{tab:metrics} lists the evaluation metric and tool for each benchmark.
GSM8K, BBH, MathQA, WinoGrande, and PIQA use exact match.
HumanEval and MBPP use pass@1.
LongBench uses F1 and ROUGE-L.

\begin{table}[t]
\centering
\small
\setlength{\tabcolsep}{3pt}
\resizebox{\columnwidth}{!}{
\begin{tabular}{lll}
\toprule
\textbf{Benchmark} & \textbf{Metric} & \textbf{Tool} \\
\midrule
GSM8K & Exact match & \texttt{lm-eval} \\
BBH & Exact match & \texttt{lm-eval} \\
HumanEval & pass@1 & \texttt{lm-eval} \\
MBPP & pass@1 & \texttt{lm-eval} \\
MathQA & Exact match & \texttt{lm-eval} \\
WinoGrande & Exact match & \texttt{lm-eval} \\
PIQA & Exact match & \texttt{lm-eval} \\
LongBench & F1 / ROUGE-L & LongBench scripts \\
\bottomrule
\end{tabular}}
\caption{
Evaluation metric and tool per benchmark.
}
\label{tab:metrics}
\end{table}

\subsection{Baselines}
\label{app:baselines}

We compare against No-cache, DualCache~\citep{wu2025fastdllm}, and ES-dLLM~\citep{zhu2026esdllm}.
No-cache is vanilla diffusion decoding without KV caching or parallel decoding.
DualCache caches KV activations for both prompt and response tokens.
ES-dLLM is run using its official implementation and published settings.
We use importance score $\alpha=0.5$ and proportion steps $[(1.0,0.0),(0.5,0.125),(0.25,0.25)]$.
No per-benchmark tuning is performed for any baseline.

FlashDLM~\citep{hu2025flashdlm}, Sparse dLLM~\citep{song2025sparsedllm}, and D2F~\citep{wang2025d2f} are not included as direct baselines because compatible public implementations for both LLaDA-8B and Dream-7B were not available at the time of evaluation.
Spiffy~\citep{agrawal2026spiffy} is orthogonal to our method because it uses speculative decoding, and a combined comparison is left for future work.

\subsection{Statistical Errors and Determinism}
\label{app:statistics}

Standard errors are computed by \texttt{lm-eval}~\citep{eval-harness} as the standard error of the mean across evaluation samples.
Table~\ref{tab:stderr} reports the standard errors for \textsc{CAI-dLLM}.
Errors for other methods are of similar magnitude.

\begin{table}[t]
\centering
\small
\setlength{\tabcolsep}{3pt}
\resizebox{\columnwidth}{!}{
\begin{tabular}{lccc}
\toprule
\textbf{Bench.} & \textbf{Dream acc.} 
  & \textbf{LLaDA acc.} & \textbf{Std.\ err.} \\
\midrule
GSM8K & 77.26\% & 77.41\% & $\pm$1.18 pp \\
BBH & 56.95\% & 52.33\% & $\pm$0.53 pp \\
MBPP & 56.60\% & 37.00\% & $\pm$2.16 pp \\
HumanEval & 48.17\% & 35.98\% & $\pm$3.76 pp \\
\bottomrule
\end{tabular}}
\caption{
\textsc{CAI-dLLM} accuracy and standard errors.
}
\label{tab:stderr}
\end{table}

The largest accuracy difference between \textsc{CAI-dLLM} and No-cache is 4.4 points on LLaDA BBH.
The standard error for this result is $\pm 0.53$, so the gap is about eight times larger than the estimated uncertainty.
This suggests that the BBH difference is unlikely to be caused only by sampling noise as BBH contains multi-step reasoning tasks, where early token commitment can sometimes preserve a wrong intermediate choice and reduce the chance of later correction.
HumanEval has higher uncertainty because it has only 164 problems, so its differences should be interpreted with more care.

Accuracy evaluation uses \texttt{lm-eval} with fixed seeds: random seed 0, NumPy seed 1234, and torch manual seed 1234.
Decoding is deterministic at temperature 0.
Timing measurements are taken after one warm-up pass.

\subsection{Software Versions}
\label{app:software}

All experiments use PyTorch 2.3.1, CUDA 12.4, Transformers 4.44.0, and Python 3.11.
The LLaDA-8B-Instruct and Dream-7B-Instruct weights are loaded from HuggingFace Hub using official checkpoints.
Flash Attention 2 is enabled for all methods.
CUDA deterministic mode is enabled for all runs.
Results may differ on earlier PyTorch versions due to different CUDA kernel availability on H200.

\subsection{Compute Budget}
\label{app:compute_budget}

All experiments use a single NVIDIA H200 GPU.
As shown in Table~\ref{tab:compute_budget}, total inference compute across all benchmarks, models, and methods is approximately 99.4 GPU-hours.
No-cache inference alone requires 77.4 hours, or 77.9\% of the total.
In comparison, \textsc{CAI-dLLM} completes the same evaluation in 3.9 hours, reducing inference compute by 95.0\% and saving 73.5 GPU-hours relative to no-cache decoding.

\begin{table}[t]
\centering
\small
\setlength{\tabcolsep}{5pt}
\begin{tabular}{lrr}
\toprule
\textbf{Method} & \textbf{Hours} & \textbf{\% Total} \\
\midrule
No cache & 77.38 & 77.9\% \\
ES-dLLM & 8.19 & 8.2\% \\
DualCache & 8.07 & 8.1\% \\
\textsc{CAI-dLLM} & 3.89 & 3.9\% \\
\textsc{CAI-dLLM} no CG & 1.85 & 1.9\% \\
\midrule
\textbf{Total} & \textbf{99.38} & 100\% \\
\bottomrule
\end{tabular}
\caption{
Inference GPU-hours by method.
CG denotes confidence gating.
}
\label{tab:compute_budget}
\end{table}

\section{Additional Ablation Results}
\label{app:ablation_results}

\subsection{GSM8K Ablation}
\label{app:gsm8k_ablation}

Table~\ref{tab:gsm8k_ablation} reports the GSM8K component ablation.
The results follow the same pattern as the HumanEval ablation in Section~\ref{sec:ablation}.
APD gives the main throughput gain for both models.
The per-block schedule gives a small additional gain.
For LLaDA, adding token budgets with confidence gating improves accuracy to 77.41\%, which is higher than ES-dLLM.
For Dream, confidence gating does not improve accuracy or speed, so the best configuration keeps gating off.

\begin{table}[t]
\centering
\scriptsize
\setlength{\tabcolsep}{2.5pt}
\renewcommand{\arraystretch}{0.92}
\begin{tabular}{llcc}
\toprule
\textbf{Model} & \textbf{Config.} & \textbf{Acc.} & \textbf{TPS} \\
\midrule
\multirow{5}{*}{LLaDA}
& ES-dLLM                        & 77.10          & 175.3 \\
& +APD                           & 76.80          & 490.2 \\
& +APD+Block                     & 76.95          & \textbf{496.0} \\
& +APD+Block+Budget $\star$      & \textbf{77.41} & 335.1 \\
& Full CAI, gate off             & 76.57          & 320.0 \\
\midrule
\multirow{5}{*}{Dream}
& ES-dLLM                        & \textbf{77.48} & 97.9  \\
& +APD                           & 75.36          & 372.9 \\
& +APD+Block                     & 75.51          & \textbf{383.0} \\
& +APD+Block+Budget              & 75.51          & \textbf{383.0} \\
& Full CAI, gate off $\star$     & 77.26          & 369.5 \\
\bottomrule
\end{tabular}
\caption{
GSM8K ablation of \textsc{CAI-dLLM} components.
$\star$ marks the recommended configuration for each model.
}
\label{tab:gsm8k_ablation}
\end{table}

\section{Additional Background and Motivation Analysis}
\label{app:background_motivation}

This appendix provides additional evidence for the observations in Section~\ref{sec:background}.
The main paper focuses on the most direct motivation: denoising behavior is uneven across layers, blocks, and tokens.
Here, we include extra analysis on attention sparsity, APD schedule sensitivity, and hardware behavior.

\paragraph{Attention sparsity across layers.}
LLaDA-8B has 32 transformer layers (L0--L31). 
We sample six layers at roughly equal intervals 
(L0, L4, L8, L16, L24, L31) to represent early, 
middle, and late processing stages without 
profiling all 32 layers at full cost.

\begin{figure*}[t]
  \centering
  \includegraphics[width=0.90\textwidth]{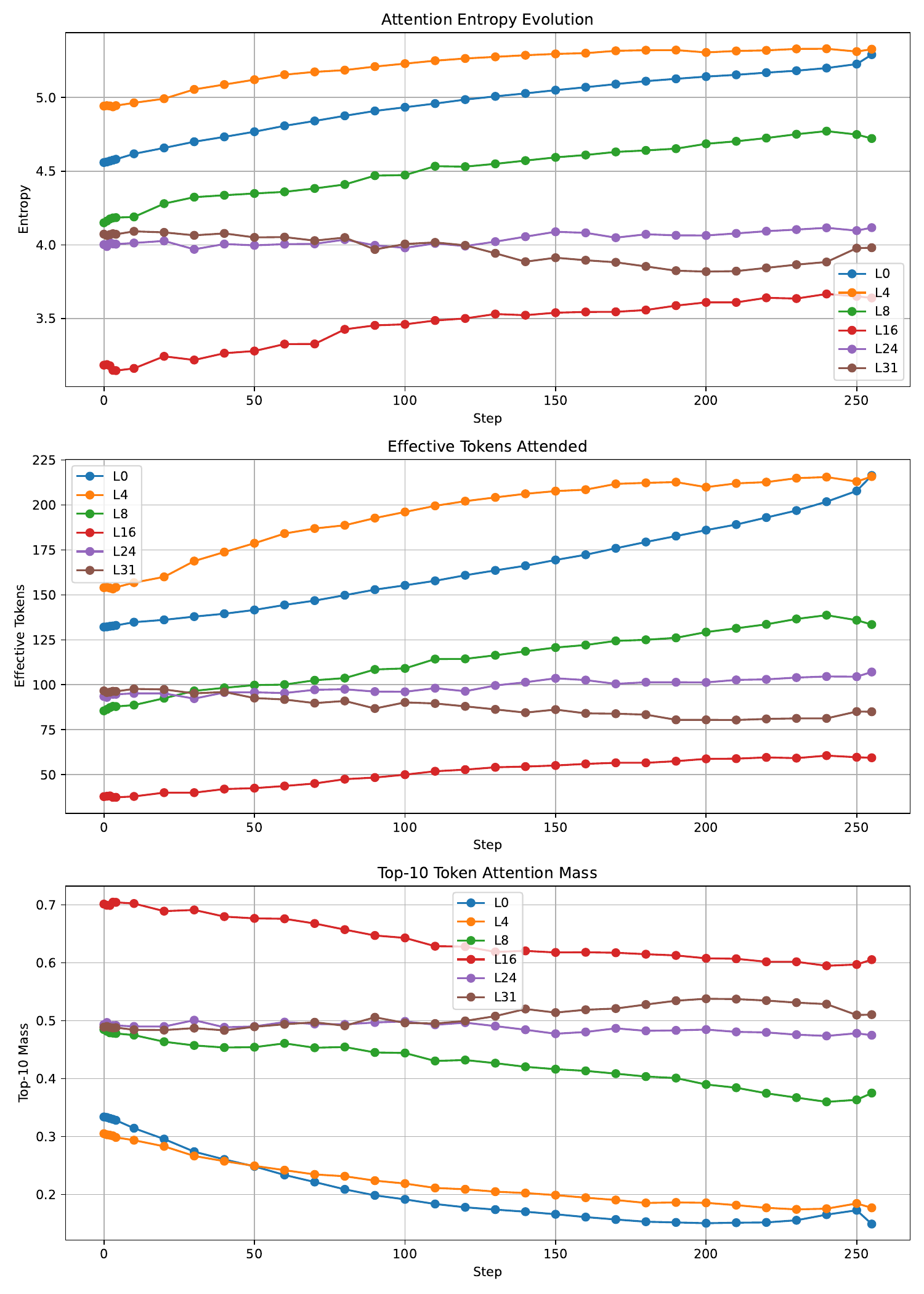}
  \caption{
  Attention sparsity across denoising steps for sampled layers of LLaDA-8B on GSM8K.
  We sample early, middle, and late transformer layers.
  The plots show attention entropy, effective tokens attended, and top-10 attention mass.
  The strong layer-to-layer variation supports layer-aware analysis of diffusion language model inference.
  }
  \label{fig:app_attention_sparsity}
\end{figure*}

Figure~\ref{fig:app_attention_sparsity} shows three 
complementary signals across these layers. First, 
attention entropy decreases in deep layers (L24, L31) 
and increases in shallow layers (L0, L4) as denoising 
progresses, showing that deeper layers become 
progressively more selective over time. Second, the 
number of effective tokens attended drops sharply in 
deep layers, confirming that later layers concentrate 
on fewer positions while early layers attend broadly. 
Third, the top-10 attention mass is highest in deep 
layers, indicating that a small set of tokens captures 
most of the attention weight in those layers.

Together, these observations show that compute 
requirements differ substantially across layers: early 
layers process all tokens broadly while deep layers 
focus selectively. A single uniform compute policy 
across all layers is therefore wasteful, which 
motivates layer-aware or token-aware inference 
strategies such as \textsc{CAI-dLLM}.

\begin{figure*}[t]
  \centering
  \includegraphics[width=0.95\textwidth]{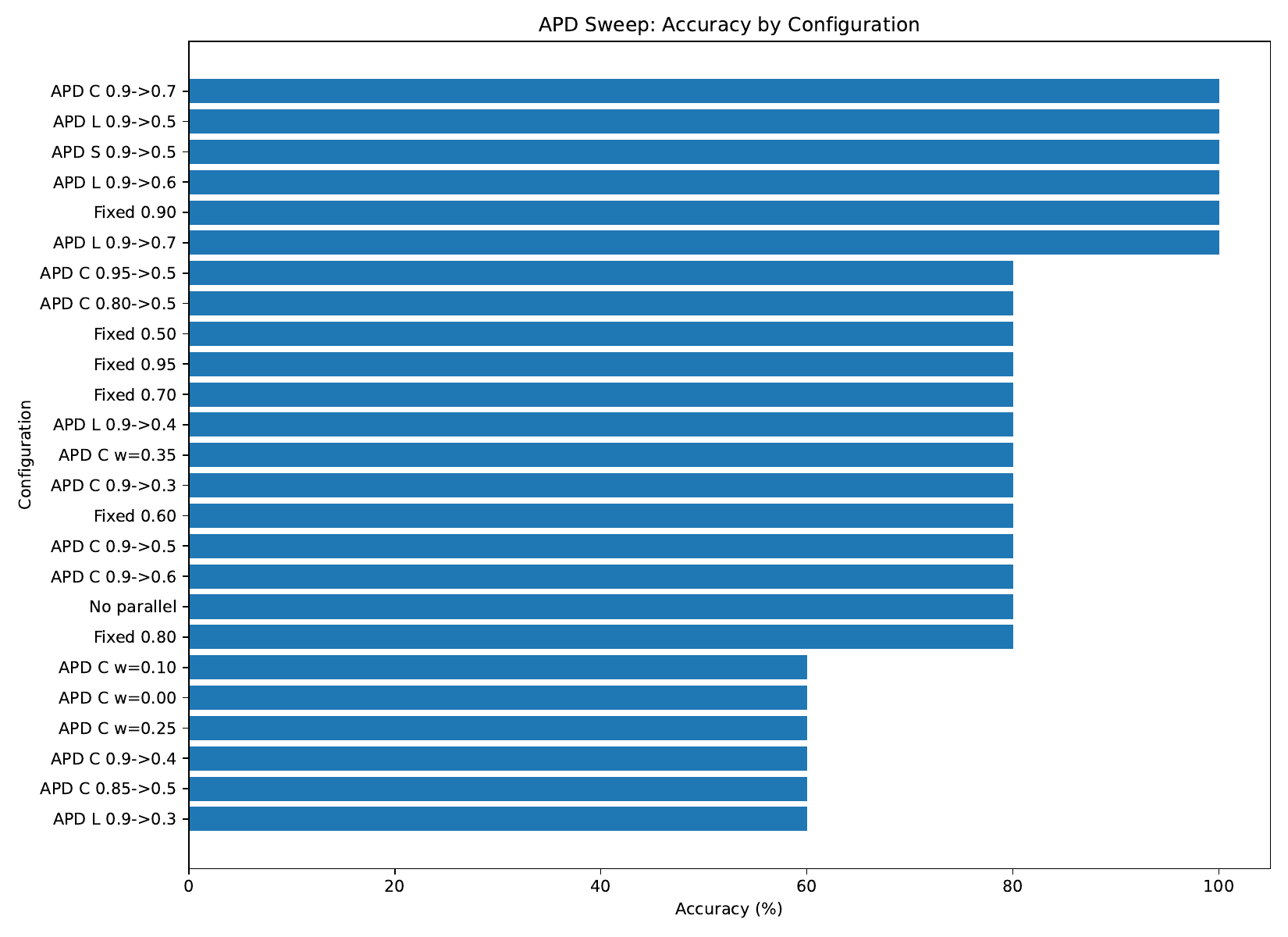}
  \caption{
  Accuracy across APD schedule configurations on LLaDA-8B GSM8K.
  The selected cosine schedule with $\theta_s=0.90$, $\theta_e=0.70$, and $w=0.15$ gives the best accuracy-speed trade-off.
  }
  \label{fig:app_apd_sweep}
\end{figure*}

\paragraph{APD schedule sensitivity.}
Figure~\ref{fig:app_apd_sweep} shows that decoding accuracy depends on the confidence schedule.
Fixed thresholds and overly aggressive schedules can reduce accuracy.
The selected cosine schedule gives the best accuracy-speed trade-off among the tested settings.
This supports using an adaptive threshold schedule rather than a single fixed threshold.

\begin{figure*}[t]
  \centering
  \includegraphics[width=\textwidth]{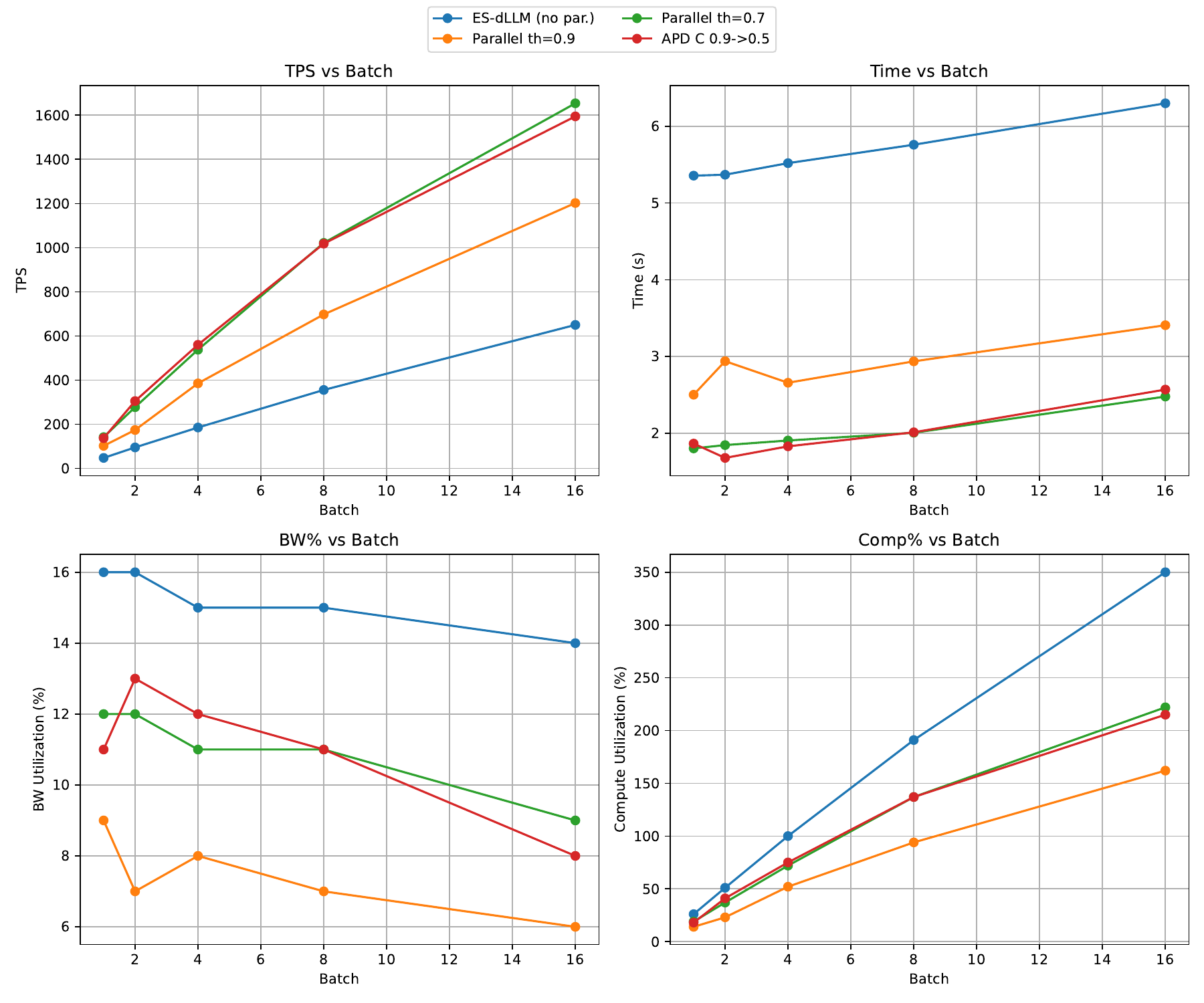}
  \caption{
  Hardware behavior across batch sizes.
  Fixed-threshold parallel decoding and APD are compared with ES-dLLM.
  The H200 runs the workload in a compute-bound regime, making denoising-step reduction an effective optimization target.
  }
  \label{fig:app_roofline}
\end{figure*}

\paragraph{Hardware behavior.}
Figure~\ref{fig:app_roofline} compares throughput, 
latency, bandwidth use, and compute use across batch 
sizes on the H200 GPU. The four curves correspond to 
different decoding configurations: \texttt{ES-dLLM 
(no par.)} is ES-dLLM without parallel decoding; 
\texttt{Parallel th=0.9} and \texttt{Parallel th=0.7} 
are confidence based parallel decoding with fixed 
commit thresholds of 0.9 and 0.7 respectively, used 
here as ablation points to understand the effect of 
threshold choice on hardware utilization; and 
\texttt{APD C 0.9$\rightarrow$0.5} is our adaptive cosine 
schedule, which starts at 0.9 and relaxes to 0.5. 
These are not separate systems but configurations 
used to profile hardware behavior under different 
decoding policies.

As batch size increases, compute utilization 
(bottom-right panel) grows much more steeply than 
bandwidth utilization (bottom-left panel) across all 
configurations. At batch size 8, compute utilization 
exceeds 150--350\% of baseline while bandwidth 
utilization stays below 16\%. This gap confirms that 
the bottleneck is compute, not memory bandwidth. On 
a compute bound workload, reducing the number of 
denoising steps directly reduces wall-clock time and 
energy, which is the primary optimization axis of 
\textsc{CAI-dLLM}.

\section{Additional Results and Analysis}
\label{app:detailed_results}

This appendix reports memory usage, task-level LongBench results, and detailed energy measurements.
These results support the main paper and show that \textsc{CAI-dLLM} improves efficiency without large memory overhead.

\subsection{Memory Usage}
\label{app:memory}

Table~\ref{tab:app_memory} reports peak GPU memory 
on HumanEval. HumanEval is used for profiling because 
its 164 problems make repeated profiling practical. 
Memory use is mainly determined by model weights and 
KV cache size, so the results are representative 
across tasks.

For LLaDA, \textsc{CAI-dLLM} uses 32,787\,MB, only 
480\,MB more than DualCache (1.5\% overhead). This 
extra memory comes from three small components: 
per-token first-step confidence scores $s_i$ 
($\mathcal{O}(L)$ floats), token step budgets $B_i$ 
($\mathcal{O}(L)$ integers), and the position aware 
threshold lookup table ($\mathcal{O}(L)$ floats). 
Together these are negligible relative to the model 
weights and KV cache.

For Dream, \textsc{CAI-dLLM} uses 36,533\,MB, which 
is 5,822\,MB more than DualCache but still 7,500\,MB 
below no-cache decoding. The larger increase is not 
caused by the confidence controller itself. Dream 
uses more aggressive parallel decoding, which fills 
a larger output token buffer simultaneously. 
Specifically, the adaptive threshold schedule commits 
more tokens in parallel per step on Dream than on 
LLaDA, increasing the size of the active output 
buffer. The \textsc{CAI-dLLM} bookkeeping overhead is the 
same $\mathcal{O}(L)$ as for LLaDA; the difference 
is entirely due to Dream's parallel decoding 
behavior.

\begin{table}[t]
\centering
\small
\setlength{\tabcolsep}{4pt}
\begin{tabular}{llcc}
\toprule
\textbf{Model} & \textbf{Method} 
  & \textbf{Peak Mem.} & \textbf{vs DualCache} \\
\midrule
LLaDA & No cache         & 41,567\,MB & +9,260\,MB \\
LLaDA & DualCache        & 32,307\,MB & baseline   \\
LLaDA & ES-dLLM          & 32,503\,MB & +196\,MB   \\
LLaDA & \textsc{CAI-dLLM} & 32,787\,MB & +480\,MB  \\
\midrule
Dream & No cache         & 44,033\,MB & +13,322\,MB \\
Dream & DualCache        & 30,711\,MB & baseline    \\
Dream & ES-dLLM          & 30,807\,MB & +96\,MB     \\
Dream & \textsc{CAI-dLLM} & 36,533\,MB & +5,822\,MB \\
\bottomrule
\end{tabular}
\caption{
Peak GPU memory on HumanEval.
The extra memory for \textsc{CAI-dLLM} over DualCache comes from per-token confidence scores, budgets, and threshold tables.
Dream's larger overhead is due to fuller parallel output generation, not \textsc{CAI-dLLM} bookkeeping.
}
\label{tab:app_memory}
\end{table}

\subsection{Detailed LongBench Results}
\label{app:longbench_results}

Table~\ref{tab:app_longbench} reports task-level 
LongBench results on Dream-7B. Each task uses its 
own metric: F1 for NarrativeQA and MultifieldQA-en, 
and ROUGE-L for GovReport and QMSum. The average 
score is the unweighted mean of these four task 
scores, where each score is on a 0--100 scale.

\textsc{CAI-dLLM} increases average throughput from 
2.7 to 51.4 tokens per second, giving a 
$19.04\times$ speedup over no-cache decoding. The 
average score decreases from 24.63 to 23.11, a drop 
of 1.52 points, which is similar to the drops seen 
for DualCache (1.10 points) and ES-dLLM (0.94 
points). This shows that all caching methods reduce 
long-context quality by a similar margin, while 
\textsc{CAI-dLLM} provides more than double their 
throughput. On QMSum, \textsc{CAI-dLLM} slightly 
improves over both cached baselines (14.66 vs.\ 
14.59 and 14.38), suggesting that for summarization 
tasks the confidence aware commit schedule does not 
hurt quality. The larger drop on MultifieldQA-en 
(43.72 vs.\ 47.95 for ES-dLLM) indicates that 
multi-field retrieval tasks are more sensitive to 
early token commitment. Overall, these results 
confirm that confidence aware decoding transfers 
to long-context inference with competitive quality 
and substantially higher throughput.

\begin{table}[t]
\centering
\scriptsize
\setlength{\tabcolsep}{2.5pt}
\resizebox{\columnwidth}{!}{
\begin{tabular}{llcccc}
\toprule
\textbf{Task} & \textbf{Metric} & \textbf{No cache} 
  & \textbf{DualCache} & \textbf{ES-dLLM} 
  & \textbf{\textsc{CAI-dLLM}} \\
\midrule
NarrativeQA     & F1      & 18.18 & 16.96 & 17.05 & 16.96 \\
MultifieldQA-en & F1      & 47.24 & 45.35 & 47.95 & 43.72 \\
GovReport       & ROUGE-L & 19.99 & 17.23 & 16.26 & 17.09 \\
QMSum           & ROUGE-L & 13.69 & 14.59 & 14.38 & 14.66 \\
\midrule
Average score   &         & 24.63 & 23.53 & 23.69 & 23.11 \\
Average TPS     &         & 2.7   & 23.4  & 24.2  & 51.4  \\
Speedup         &         & $1.00\times$ & $8.7\times$ 
                          & $9.0\times$ & $19.04\times$ \\
\bottomrule
\end{tabular}}
\caption{
Detailed LongBench results on Dream-7B.
Average score is the unweighted mean of per-task scores.
}
\label{tab:app_longbench}
\end{table}

\subsection{Short Commonsense Tasks}
\label{app:short_commonsense}

On PIQA and WinoGrande, \textsc{CAI-DLLM} improves throughput but loses accuracy compared with ES-dLLM.
These tasks have short outputs, so there are fewer redundant denoising steps to remove.
In this setting, early commitment can force incorrect tokens before enough context is available.
Thus, \textsc{CAI-DLLM} is less suitable for short, quality-critical tasks where the small speedup does not justify the accuracy cost.

\subsection{Energy Measurements}
\label{app:energy}

Energy is computed as $E=\bar{P}\times T$, where $\bar{P}$ is the average GPU power and $T$ is wall-clock generation time.
We report energy in both Joules and Watt-hours, with $\mathrm{Wh}=E/3600$.
GPU power is sampled using \texttt{nvidia-smi}.
These measurements include GPU power only and do not include CPU, DRAM, or system level power.

Table~\ref{tab:app_energy} shows that \textsc{CAI-dLLM} reduces total energy in all measured cases.
The main reason is shorter generation time.
For example, LLaDA GSM8K drops from 3559.1 Wh to 168.7 Wh, a 95.3\% reduction.

\begin{table*}[!t]
\centering
\small
\setlength{\tabcolsep}{4pt}
\resizebox{\textwidth}{!}{
\begin{tabular}{ll l c c c c}
\toprule
\textbf{Task} & \textbf{Model} & \textbf{Method}
& \textbf{Avg Power} & \textbf{Energy (J)} & \textbf{Energy (Wh)} & \textbf{Saved} \\
\midrule
GSM8K & LLaDA & No cache & 688.0 W & 12,812,627 & 3559.1 & -- \\
GSM8K & LLaDA & DualCache & 587.2 W & 950,104 & 263.9 & 92.6\% \\
GSM8K & LLaDA & ES-dLLM & 441.7 W & 801,455 & 222.6 & 93.7\% \\
GSM8K & LLaDA & \textsc{CAI-dLLM} & 558.3 W & 607,297 & 168.7 & 95.3\% \\
\midrule
HumanEval & Dream & No cache & 673.4 W & 1,347,290 & 374.25 & -- \\
HumanEval & Dream & DualCache & 550.5 W & 215,714 & 59.92 & 84.0\% \\
HumanEval & Dream & ES-dLLM & 358.9 W & 149,309 & 41.47 & 88.9\% \\
HumanEval & Dream & \textsc{CAI-dLLM} & 464.1 W & 85,134 & 23.65 & 93.7\% \\
\midrule
HumanEval & LLaDA & No cache & 672.2 W & 1,530,033 & 425.01 & -- \\
HumanEval & LLaDA & DualCache & 529.9 W & 197,734 & 54.93 & 87.1\% \\
HumanEval & LLaDA & ES-dLLM & 348.6 W & 152,845 & 42.46 & 90.0\% \\
HumanEval & LLaDA & \textsc{CAI-dLLM} & 474.0 W & 118,921 & 33.03 & 92.2\% \\
\midrule
MBPP & Dream & No cache & 678.3 W & 3,123,006 & 867.5 & -- \\
MBPP & Dream & DualCache & 380.0 W & 375,572 & 104.3 & 88.0\% \\
MBPP & Dream & ES-dLLM & 378.1 W & 281,825 & 78.3 & 91.0\% \\
MBPP & Dream & \textsc{CAI-dLLM} & 412.6 W & 122,798 & 34.1 & 96.1\% \\
\midrule
MBPP & LLaDA & No cache & 676.3 W & 3,704,466 & 1029.0 & -- \\
MBPP & LLaDA & DualCache & 413.4 W & 349,628 & 97.1 & 90.6\% \\
MBPP & LLaDA & ES-dLLM & 371.7 W & 291,715 & 81.0 & 92.1\% \\
MBPP & LLaDA & \textsc{CAI-dLLM} & 434.4 W & 180,389 & 50.1 & 95.1\% \\
\bottomrule
\end{tabular}}
\caption{
Detailed energy measurements.
The Saved column reports reduction relative to the no-cache row for the same task and model.
}
\label{tab:app_energy}
\end{table*}

\section{Throughput Benchmark Results}
\label{app:throughput}

We report wall-clock throughput in tokens per second (TPS) and speedup over the \texttt{nocache} baseline.
We evaluate four generation lengths, $L \in \{64, 128, 256, 512\}$, on two instruction tuned diffusion LLMs: LLaDA-Instruct~\cite{nie2025llada} and Dream-Instruct~\cite{ye2025dream}.
All experiments use a single NVIDIA H200 GPU under the same hardware and software setup.
For each generation length, speedup is computed as the ratio between a method's TPS and the \texttt{nocache} TPS.

\begin{table*}[t]
\centering
\scriptsize
\setlength{\tabcolsep}{3pt}
\resizebox{\textwidth}{!}{
\begin{tabular}{llcccccccc}
\toprule
\textbf{Model} & \textbf{Method}
& \multicolumn{2}{c}{$L=64$}
& \multicolumn{2}{c}{$L=128$}
& \multicolumn{2}{c}{$L=256$}
& \multicolumn{2}{c}{$L=512$} \\
\cmidrule(lr){3-4}
\cmidrule(lr){5-6}
\cmidrule(lr){7-8}
\cmidrule(lr){9-10}
& & TPS & Speedup & TPS & Speedup & TPS & Speedup & TPS & Speedup \\
\midrule
\multirow{4}{*}{LLaDA}
& No cache & 52.3 & 1.00$\times$ & 53.8 & 1.00$\times$ & 54.0 & 1.00$\times$ & 40.6 & 1.00$\times$ \\
& DualCache & 51.3 & 0.98$\times$ & 50.2 & 0.93$\times$ & 50.8 & 0.94$\times$ & 51.4 & 1.27$\times$ \\
& ES-dLLM & 45.4 & 0.87$\times$ & 46.9 & 0.87$\times$ & 46.8 & 0.87$\times$ & 46.9 & 1.16$\times$ \\
& \textsc{CAI-dLLM} & \textbf{108.0} & \textbf{2.07$\times$} & \textbf{202.1} & \textbf{3.76$\times$} & \textbf{123.6} & \textbf{2.29$\times$} & \textbf{186.3} & \textbf{4.59$\times$} \\
\midrule
\multirow{4}{*}{Dream}
& No cache & 61.5 & 1.00$\times$ & 63.1 & 1.00$\times$ & 61.4 & 1.00$\times$ & 45.3 & 1.00$\times$ \\
& DualCache & 62.4 & 1.01$\times$ & 63.5 & 1.01$\times$ & 64.3 & 1.05$\times$ & 64.0 & 1.41$\times$ \\
& ES-dLLM & 57.5 & 0.93$\times$ & 58.7 & 0.93$\times$ & 59.7 & 0.97$\times$ & 59.6 & 1.32$\times$ \\
& \textsc{CAI-dLLM} & \textbf{98.6} & \textbf{1.60$\times$} & \textbf{199.2} & \textbf{3.16$\times$} & \textbf{242.9} & \textbf{3.96$\times$} & \textbf{692.6} & \textbf{15.29$\times$} \\
\bottomrule
\end{tabular}}
\caption{
Throughput and speedup over No cache across generation lengths on LLaDA-Instruct and Dream-Instruct.
Best TPS for each generation length is shown in bold.
}
\label{tab:tps_combined}
\end{table*}

As shown in Table~\ref{tab:tps_combined}, \textsc{CAI-dLLM} achieves the highest throughput in every setting across both models.
On LLaDA-Instruct, speedup ranges from 2.07$\times$ at $L=64$ to 4.59$\times$ at $L=512$.
This shows that longer generation lengths give more room for adaptive step reduction.
On Dream-Instruct, the gain is larger and reaches 15.29$\times$ over \texttt{nocache} at $L=512$ (692.6 vs.\ 45.3 TPS).
In contrast, \texttt{dualcache} and \texttt{esdllm} do not improve throughput at short generation lengths and sometimes slightly reduce it.
This happens because their fixed caching overheads are not fully amortized when $L \in \{64,128\}$.
Overall, these results show that \textsc{CAI-dLLM} is most effective at longer sequence lengths, where token difficulty varies more and early commitment saves more computation.

\end{document}